\documentclass[lettersize,journal]{IEEEtran}
\usepackage{amsmath,amsfonts}
\usepackage{algorithmic}
\usepackage{array}
\usepackage[caption=false,font=normalsize,labelfont=sf,textfont=sf]{subfig}
\usepackage{textcomp}
\usepackage{stfloats}
\usepackage{url}
\usepackage{gensymb}
\usepackage{verbatim}
\pdfoutput=1
\usepackage{adjustbox}
\usepackage{graphicx}
\usepackage{multirow}
\def\BibTeX{{\rm B\kern-.05em{\sc i\kern-.025em b}\kern-.08em
    T\kern-.1667em\lower.7ex\hbox{E}\kern-.125emX}}
\usepackage{balance}
\begin{document}
%\title{Differentiable SAR Renderer Embedded Reinforcement Learning for View Angle Inversion in SAR images}
%\title{Reinforcement Learning for SAR Target Orientation Inference with Differentiable SAR Renderer}
\title{Reinforcement Learning for SAR View Angle Inversion with Differentiable SAR Renderer}
% {Reinforcement Learning for SAR Images View Angles Inversion Based on DSR}
\author{Yanni Wang, \textit{Graduate Student Member, IEEE}, Hecheng Jia, \textit{Graduate Student Member, IEEE}, \\Shilei Fu, \textit{Graduate Student Member, IEEE}, Huiping Lin, \textit{Member, IEEE}, Feng Xu, \textit{Senior Member, IEEE}

\thanks{This work was supported in part by the National Natural Science Foundation of China under Grant U2130202, and in part by the Fundamental Research Funds for the Central Universities. \textit{(Corresponding Authors: Huiping Lin, Feng Xu)}.}% <-this % stops a space

\thanks{The authors are with the Key Laboratory for Information Science of
Electromagnetic Waves (MoE), Fudan University, Shanghai 200433, China. (email: fengxu@fudan.edu.cn)}
}

%\markboth{Journal of \LaTeX\ Class Files,~Vol.~18, No.~9, September~2020}
\markboth{}
{Differentiable SAR Renderer Embedded Reinforcement Learning for View Angle Inversion in SAR images}
\maketitle

\begin{abstract}
The electromagnetic inverse problem has long been a research hotspot. This study aims to reverse radar view angles in synthetic aperture radar (SAR) images given a target model. Nonetheless, the scarcity of SAR data, combined with the intricate background interference and imaging mechanisms, limit the applications of existing learning-based approaches. To address these challenges, we propose an interactive deep reinforcement learning (DRL) framework, where an electromagnetic simulator named differentiable SAR render (DSR) is embedded to facilitate the interaction between the agent and the environment, simulating a human-like process of angle prediction. Specifically, DSR generates SAR images at arbitrary view angles in real time. And the differences in sequential and semantic aspects between the view angle-corresponding images are leveraged to construct the state space in DRL, which effectively suppress the complex background interference, enhance the sensitivity to temporal variations, and improve the capability to capture fine-grained information. Additionally, in order to maintain the stability and convergence of our method, a series of reward mechanisms, such as memory difference, smoothing and boundary penalty, are utilized to form the final reward function. Extensive experiments performed on both simulated and real datasets demonstrate the effectiveness and robustness of our proposed method. When utilized in the cross-domain area, the proposed method greatly mitigates inconsistency between simulated and real domains, outperforming reference methods significantly.

%The electromagnetic inverse problem has long been a central focus of research. This study aims to accurately determine radar view angles in synthetic aperture radar (SAR) images using a given target model. However, the limited availability of SAR data, along with complex background noise and sophisticated imaging mechanisms, constrains the effectiveness of current learning-based methods. To overcome these challenges, we introduce an interactive deep reinforcement learning (DRL) framework that incorporates a novel electromagnetic simulator, named differentiable SAR render (DSR), to enhance the agent-environment interaction, emulating a human-like angle prediction process. DSR is capable of generating SAR images from various view angles instantaneously. It utilizes discrepancies in sequential and semantic information between images corresponding to different view angles to construct a refined state space in DRL. This space effectively mitigates background interference, increases sensitivity to temporal changes, and augments the ability to capture detailed information. Furthermore, to ensure the stability and convergence of our approach, we employ a sophisticated reward function incorporating mechanisms such as memory differentiation, smoothing, and boundary penalties. Our extensive testing on both simulated and real datasets validates the proposed method's efficiency and robustness. Notably, when applied to cross-domain scenarios, our approach significantly reduces the discrepancies between simulated and real-world data, outperforming standard methods substantially.

\end{abstract}


\section{Introduction}
\IEEEPARstart{E}{lectromagnetic} inverse problem focuses on inferring the properties of the target and the sensor's observational characteristics, given the known attributes of the target. Synthetic aperture radar (SAR) functions as an active remote sensing sensor that operates within the microwave frequency spectrum, boasting capabilities for round-the-clock and all-weather operation \cite{moreira2013tutorial}. The inversion of radar observation angles in SAR imagery remains a central electromagnetic inverse challenge. This task aims to ascertain the radar view angles of targets within the radar coordinate system from a supplied SAR image, which has critical implications for fields ranging from radar imaging and monitoring \cite{zhou2023sidelobe} to intelligent transportation \cite{ulrich2022improved} and environmental monitoring \cite{he2016soil}.
However, various imaging factors like multiple scattering phenomena, intricate background disturbances, shadowing effects and speckle noise hinder the view angle inversion process \cite{zhang2021coherent}. 
 % Notably, azimuth variations correlate with the target's non-linear rotational representation in spatial domains, posing challenges to discerning the intrinsic physical attributes of the target  \cite{song2021learning}.

\begin{figure}[t]
	\centering
		\includegraphics[width = \linewidth]{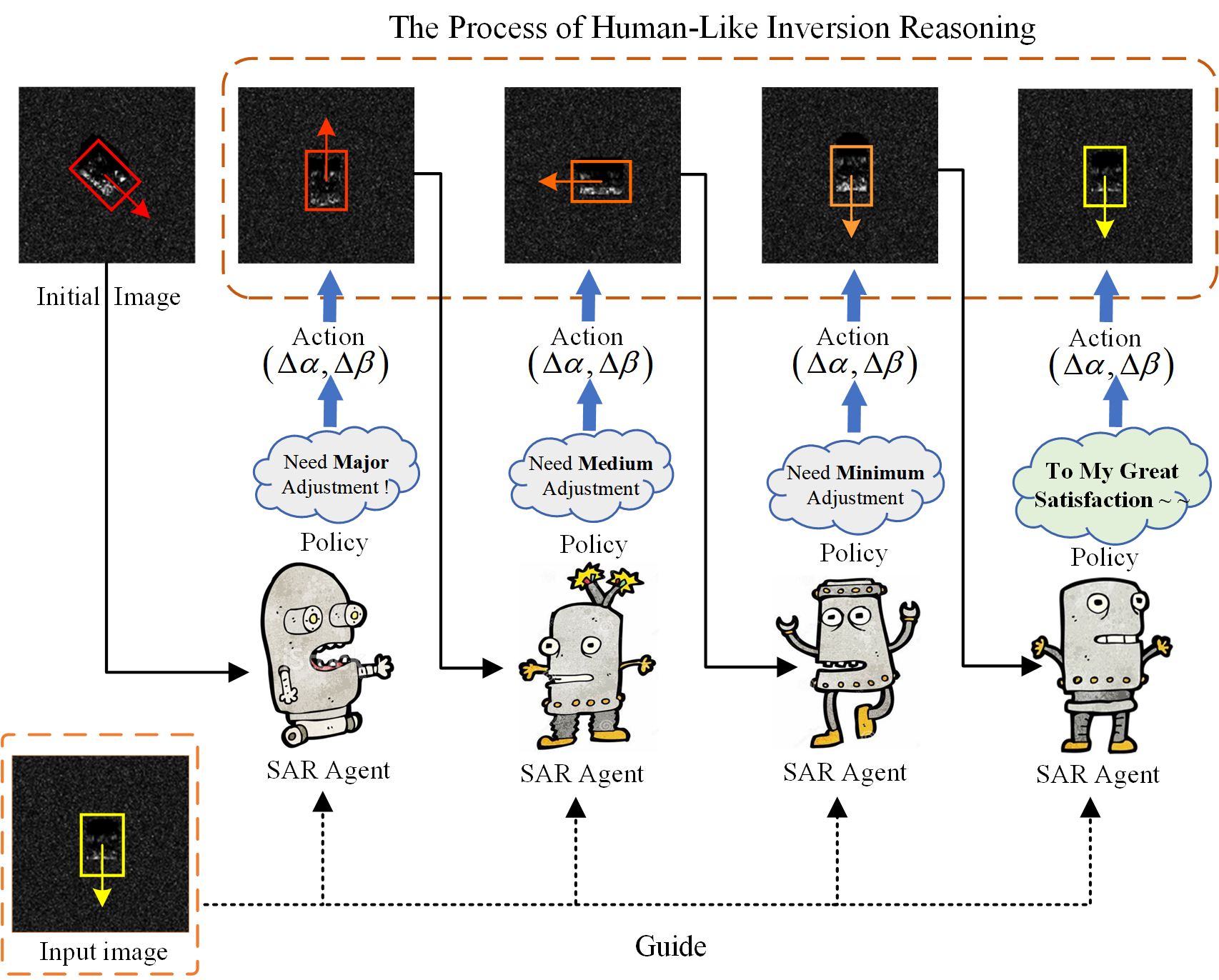}
		\caption{The SAR agent learns inverse policies and takes actions to reverse view angles of SAR image from any arbitrary, mirroring the decision-making process similar to human prediction. Due to the diverse range of view angles, the SAR agents are required to employ actions with varying adjustments based on the disparity between the reference and current inversion SAR image.}
  \vspace{-0.4cm}
\label{0}\end{figure}

% Most existing methods \cite{ruder2016overview, sutton1986two, kingma2014adam} 
% adopted traditional optimization techniques to address inverse problems, encompassing both gradient-based optimization algorithms and heuristic optimization algorithms. Gradient-based strategies cater specifically to problems with gradient information and finite solution space. 
% When the solution space is a discrete domain with high dimensions, the demand for strong robustness becomes more essential. 
% Heuristic optimization algorithms, such as the genetic algorithm (GA) \cite{holland1992adaptation} and particle swarm optimization (PSO) \cite{kennedy1995particle}, are applied to improve the robustness. Nevertheless, they are computationally intensive and tend to converge to local optima, especially when wrestling with non-convex objectives \cite{beheshti2013review}. 
% Some alternative strategies attempt to solve inverse problems by artificially constructing the inverse mapping relationship corresponding to the forward problem \cite{karakucs2019ship, kallestad2023general, zhou2022saf}. The challenge of establishing this inverse mapping relationship is due to the complexity of the forward problem itself and human factors. 

Most existing methods \cite{ruder2016overview, sutton1986two, kingma2014adam} utilize traditional optimization techniques to address inverse problems, incorporating both gradient-based and heuristic optimization algorithms. Gradient-based methods are particularly suited to problems with available gradient information and a finite solution space. However, in discrete domains with high dimensions, the necessity for robustness intensifies. Heuristic algorithms, such as genetic algorithm (GA) \cite{holland1992adaptation} and particle swarm optimization (PSO) \cite{kennedy1995particle}, enhance robustness but are computationally demanding and prone to local optima in non-convex objectives \cite{beheshti2013review}. Alternative approaches address inverse problems by constructing inverse mappings corresponding to forward problems \cite{karakucs2019ship, kallestad2023general, zhou2022saf}. The complexity of these forward problems and human factors contribute to the challenges of establishing accurate inverse mappings.

In order to overcome the limitations of traditional methods, an increasing amount of research has focused on utilizing machine learning techniques to address inverse problems. The main approach involves learning features and patterns from extensive data, which allows for the capture of complex nonlinear inverse mapping relationships. Particularly, deep learning (DL) has provided possibilities to efficiently solve inverse problems by leveraging data-driven neural networks to 
model nonlinear inverse mapping functions \cite{li2023multi, zhou2023replay}. As for SAR images' view angle inversion, Guo \emph{et al.} \cite{guo2023causal} showed that the azimuth angle prediction error varies between ${{13}^ \circ }$ and ${{33}^ \circ }$ with a DL method. Song \emph{et al.}  \cite{song2021learning} used angle inversion accuracy as an evaluation index to measure the effect of the generated model. 
Due to the costly and intricate characteristics of obtaining SAR images, coupled with a scarcity of publicly accessible SAR datasets, training on limited data could not adequately capture the vast variations and specific diversities to different targets and random viewpoints \cite{guo2022learning}. Additionally, the inverse mapping relationship established by the network does not directly utilize the actual physical process of electromagnetic scattering. Such constraints result in poor performance of learning-based networks particularly when handling unseen data \cite{feng2017deep, zhu2017deep, guo2022recognition}.

Deep reinforcement learning (DRL) has recently been recognized as a potent strategy for addressing inverse optimization problems \cite{sridharan2022deep, mazyavkina2021reinforcement, mankowitz2023faster}. DRL combines the perceptive ability of DL and the decision-making ability of reinforcement learning (RL) within interactive environments to establish the optimal policy by applying a Markov decision process and the Bellman equation \cite{mnih2015human}. Specifically, DRL guides the learning of the agent to make intelligent decisions in interactive environments by employing reinforcement signals (rewards or penalties) as feedback \cite{sutton2018reinforcement}. In comparison to traditional supervised learning methods that require massive labeled data, DRL can rely less on pre-provided strong label data to autonomously learn through interaction with the environment. Combining the ability of perceptive and decision-making allows DRL to show strong performance for complex problems, enabling learning and adaptation across different tasks \cite{mnih2015human, mnih2013playing}. Moreover, the interactive learning mode of reinforcement learning makes it a great advantage in processing cross-domain data

In the field of SAR, DRL exhibits characteristics like general artificial intelligence by employing a strong interpretability policy of interactive exploration and learning, which are optimized using feedback from the environment \cite{9952926, xu2021hyperparameter, viros2020scheduling}. For example, Xu \emph{et al.} \cite{xu2021hyperparameter} presented the reinforcement learning and hyperband (RLH) to dynamically learn hyper-parameters for ship detection from SAR images. Martin \emph{et al.} \cite{viros2020scheduling} developed a method to effectively adapt the baseline of an SAR based on DRL in distributed Earth observation missions. In addition, DRL fully utilizes temporal correlations to address optimization problems of arbitrary dimensions. However, these works only focus on a few tasks, without including the SAR imaging inverse problem. How to train human-like agents capturing global discriminative and local detailed features in SAR images to interactively address radar view angle inversion problems in cases of limited data with strong cross-domain distribution backgrounds is an interesting topic, which, unfortunately, still remains largely unknown. 

%In this paper, we propose a DRL-based framework to train human-like agents to understand electromagnetic mechanisms of the physical world using an electromagnetic simulator, which enables the physical interpretability of inverse problem-solving. As shown in Fig. \ref{0}, according to the learned strategy, the trained agent takes actions from large to small amplitude from any initial Angle to predict the truth angle. Specifically, we first introduce an embedded electromagnetic simulator named differentiable SAR render (DSR) to render SAR images at arbitrary viewing angles in real-time, which interacts with the agent to simulate the human-like process of predicting angles. In this way, we utilize continuous interactive adaptive iteration between agents and DSR to improve the interactive perception and understanding ability of agents for target angles. Then, we employ the differential features in sequential and semantic among angle-associated images to construct a state space, enabling the agent to accurately comprehend the temporal contextual information. Simultaneously, a pre-defined discrete action space is designed by incorporating dynamically changing step lengths, which allows the agent to conduct efficient searching and decision-making within the solution space. Finally, we combine a series of reward mechanisms, including memory difference, smoothing, boundary penalty and initialization noise suppression to construct a comprehensive reward function, which effectively guides the learning process of the agent toward the anticipated objectives.

In this work, we introduce a DRL-based framework for training agents to interpret electromagnetic phenomena in the physical world via an electromagnetic simulator. This approach enhances the physical interpretability of solving inverse problems. As depicted in Fig. \ref{0}, the trained agents employ strategies to progressively refine their predictions from large to small amplitude angles, converging on the true angle. Initially, we integrate a differentiable SAR render (DSR) — an embedded electromagnetic simulator — to generate SAR images at various viewing angles dynamically. This interaction between agents and DSR is pivotal in mimicking the human predictive process of angles, facilitating a continuous, adaptive learning cycle.

Further, the agents' state space is constructed using differential features in sequential and semantic aspects of angle-correlated images, which empowers the agents with an enhanced understanding of temporal context. A pre-defined discrete action space with variable step lengths is also implemented, enabling efficient and dynamic exploration and decision-making within the solution space. Finally, our reward function amalgamates several mechanisms including memory difference, smoothing, boundary penalty, and initialization noise suppression, to effectively steer the agents' learning trajectory towards achieving accurate angle prediction.

The contributions of this paper are as follows:
\begin{itemize}

\item We present a DRL-based inversion algorithm and first introduce an electromagnetic simulator embedded in the environment that interacts with the agent to simulate the human-like process of angle prediction. To our knowledge, it is the first DRL-based approach that emulates human-like behavior for solving view angle inversion problems in the SAR domain.

\item In DRL, the state space is built using variations in features of angle-corresponding images, reducing background complexity and increasing sensitivity to time-related changes. The reward function is designed with techniques like reward memory difference and smoothing, noise suppression, and boundary penalties to improve the stability and efficiency of the inversion process.

\item Extensive experimental results demonstrate the remarkable effectiveness and robustness of our proposed method in accurately predicting viewing angles of SAR images. When utilized in the cross-domain area, the proposed method significantly mitigates inconsistency between simulated and real domains.
\end{itemize}

The remainder of this article is organized as follows. Section II reviews the DSR imaging. Section III introduce the proposed method in detail. Section IV reports the experimental results. And finally, we conclude our work in Section V.

\begin{figure}[t]
	\centering
		\includegraphics[width = \linewidth]{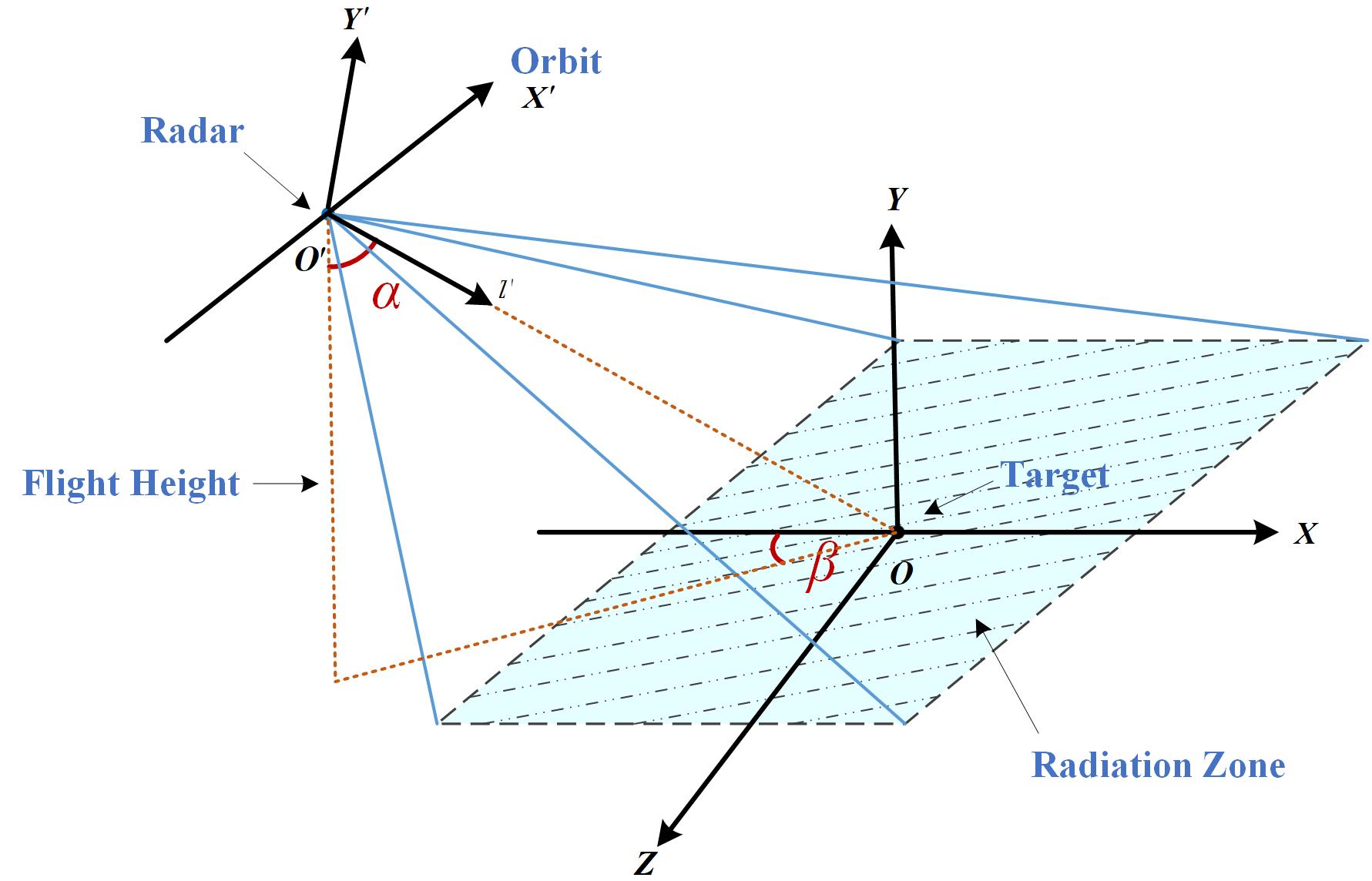}  
		\caption{SAR imaging geometric model. Related word coordinate system  \textit{O-XYZ} and the radar coordinate system $\textit{O}^{\prime}-X^{\prime}Y^{\prime}Z^{\prime}$ are defined. Our objective is to reverse the incidence angle $\alpha$ and azimuth angle $\beta$ of radar in $\textit{O}^{\prime}$-$X^{\prime}Y^{\prime}Z^{\prime}$.}
\label{coordi}\end{figure}

\begin{figure}[t]
	\centering
		\includegraphics[width = \linewidth]{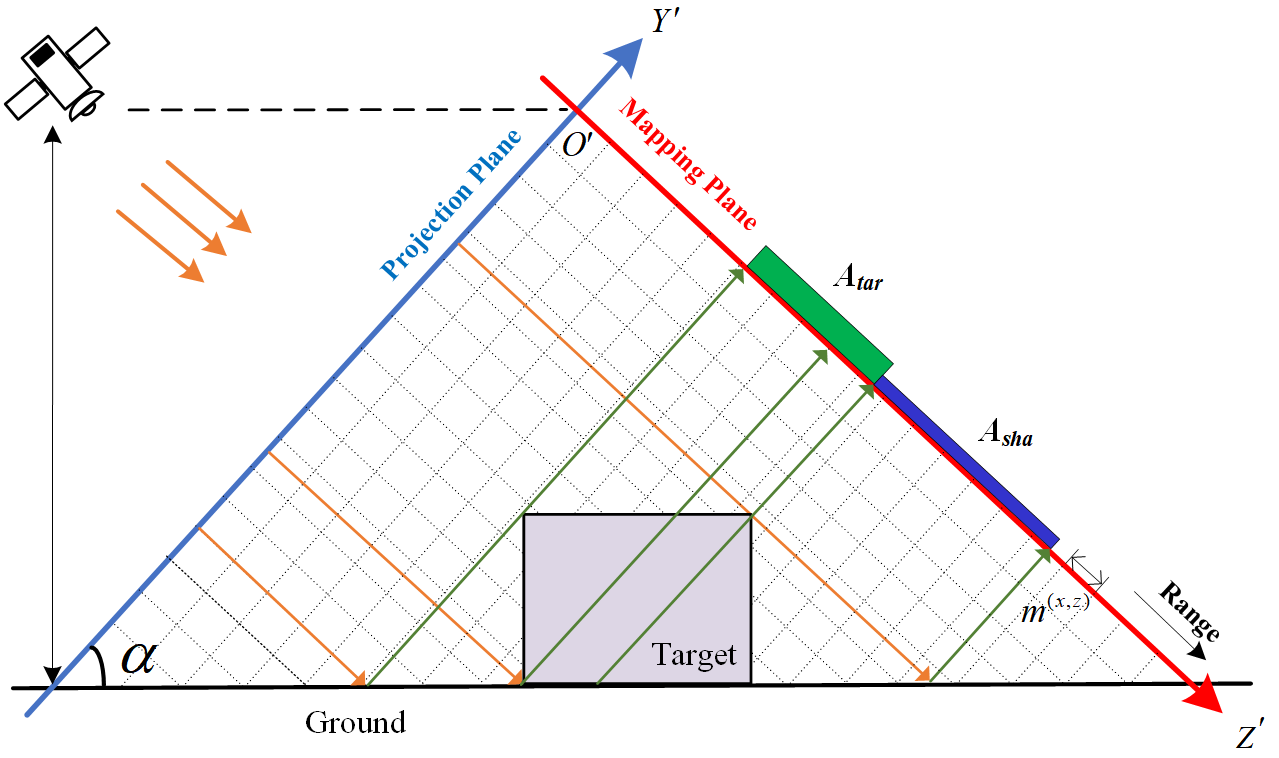}
		\caption{The mapping and projection algorithm.}
\label{mpa}\end{figure}

\section{DSR Imaging}

% SAR electromagnetic simulators are commonly used to generate SAR images by simulating environments, offering effective means to tackle real-world challenges and alleviating the constraint of limited SAR observation images. Traditional coherent echo simulation methods usually established coherent models of scatterers or computed the radar cross section (RCS) of large complex targets on rough backgrounds, such as \cite{xu2009bidirectional, yue2021coherent, zhang2016reliable}. For non-coherent simulators, Fu \emph{et al.} proposed the DSR \cite{fu2022differentiable} based on the principles of computational graphics, which generates SAR images by rendering the target from 3D to 2D based on mapping projection algorithm \cite{xu2009bidirectional}. 

% Compared to traditional coherent methods, DSR offers advantages such as ease of calling and operation, flexible invocation and fast rendering speed. Aiming to promote the agent’s exploring of inherent inversion strategy, DSR is utilized in our work to render the target from arbitrary viewpoints.

SAR electromagnetic simulators are widely employed to generate SAR images by simulating various environments, providing an effective method for addressing real-world challenges and mitigating the limitations of scarce SAR observational data. Traditional coherent echo simulation methods typically established coherent models for scatterers or computed the radar cross section (RCS) of large complex targets against rough backgrounds, as referenced in \cite{xu2009bidirectional, yue2021coherent, zhang2016reliable}. In contrast, for non-coherent simulators, Fu \emph{et al.} introduced the Differentiable SAR Renderer (DSR) \cite{fu2022differentiable}, utilizing computational graphics principles to render SAR images from 3D models to 2D projections \cite{xu2009bidirectional}.

DSR, compared to its traditional counterparts, provides several advantages including user-friendly interfacing, flexible operation, and rapid rendering capabilities. To enhance the agent’s capability in exploring inherent inversion strategies, DSR has been employed in our research to render targets from arbitrary viewpoints.

\subsection{Radar Coordinate System Definition}

% \subsubsection{Radar coordinate System definition}
In SAR imaging, the vertex coordinates of the 3D mesh shape $\mathcal{S}$ from the world coordinate system need to be converted to the radar coordinate system representation $\mathcal{S}{\prime}$. 
% Architecture.jpg

As illustrated in Fig. \ref{coordi}, the world coordinate system is denoted as \textit{O-XYZ}. In \textit{O-XYZ}, the geometric target resides at point \textit{O} and the radar 
antenna's position is \( A_p \), which is also represented as $\textit{O}^{\prime}$. Given that the radar’s azimuth orientation is $\textit{O}^{\prime}X^{\prime}$ and its slant range direction aligns with $\textit{O}^{\prime}Z^{\prime}$, the third axis can be deduced as $\textit{O}^{\prime}Y^{\prime}$. Consequently, the radar coordination system is defined as $\textit{O}^{\prime}$-$X^{\prime}Y^{\prime}Z^{\prime}$, where the incidence angle $\alpha$ refers to the angle between $\textit{O}\textit{O}^{\prime}$ and the vertical direction of the radar. The azimuth angle $\beta$ is defined as the angle between $\textit{O}\textit{O}^{\prime}$ and the reference direction.
The range of $\alpha$ is $ [ - {{90}^ \circ },{{90}^ \circ }] $, while the range of $\beta$ is $ [  {{0}^ \circ },{{360}^ \circ }]$. The radiation zone is the imaging area corresponding to the precise SAR image which is determined by parameters such as the target, the flight height and the radar perspectives $(\alpha,\beta)$. 
 
% Each triangular element has a texture value \( S \) representing its scattering intensity. The polarization mode considered in this paper is single polarization rendering, so \( S \) is a scalar. 
Then, the rotation matrix \( R \) is constructed to perform coordinate transformation from $\mathcal{S}$ to $\mathcal{S}{\prime}$ based on the defined geometric model. Specifically,
the vertex set $v_r$ in $\mathcal{S}{\prime}$ is transformed from the vertex set \( v \) in $\mathcal{S}$, which can be expressed as: 
\begin{align}
{R} = \left[ {\begin{array}{*{20}{c}}
{ - \cos \beta }&{ - \cos \alpha \sin \beta }&{ - \sin \alpha \sin \beta }\\
0&{\sin \alpha }&{ - \cos \alpha }\\
{\sin \beta }&{ - \cos \alpha \cos \beta }&{ - \sin \alpha \cos \beta }
\end{array}} \right],
\label{eqr0}
\end{align}
 \begin{align}
v_r=R^{T}(v-A_p).
\label{eqr-1}
\end{align}

\subsection{Imaging Mechanism}

The imaging mechanism in DSR is a differentiable mapping projection algorithm \cite{xu2009bidirectional, fu2022differentiable}, which contains two steps: mapping and projection.
As shown in Fig. \ref{mpa}, the imaging process is a slant-range imaging algorithm in the far field, yielding SAR images with a target area $A_{tar}$ and a shadow area $A_{sha}$. The red represents the mapping plane, parallel to the range direction; the blue signifies the projection plane, perpendicular to the mapping plane, with $O^{\prime}X^{\prime}$ as the azimuth direction.
Given the view angles (\(\alpha, \beta\)), assuming that the mapping plane contains $N_m$ mapping units, the projection plane contains $N_p$ projection units, and their corresponding relationship is shown in \ref{con:inventoryflow}:
 \begin{align}
N_p = N_m*tan(\alpha).
\label{con:inventoryflow}
\end{align}

During projection, projection units on the projection plane emit multiple rays that intersect with facets of the target in space. After striking the target, the rays scatter backward and reach the mapping plane. 
% Each unit on the mapping plane collects the scattered intensity from various facets. 
In this way, the target’s attributes from a 3D spatial are rendered onto a 2D image plane and form the $A_{tar}$. The accumulated backscattering intensity $I_{sar}^{(x,z)}$ of the target in the mapping unit $m^{(x,z)}$ is represented as \ref{con}. 
 \begin{align}
I_{\mathrm{sar}}^{(x, z)}=\sum_{n=1}^{N_f} \delta_n^{(x, z)} \cdot \omega_n^{(x, z)} \cdot S_n,
\label{con}
\end{align}
where $I_{\mathrm{sar}}^{(x, z)}$ represents the intensity of the mapping unit $m^{(x, z)}$, $N_{f}$ is the total number of target's grid facets, $\delta_n^{(x, z)}$ denotes the probability of the radar beam intersecting with facet $f_n$, $\omega_n^{(x, z)}$ is the ray intensity of $f_n$ on $m^{(x, z)}$, and $S_n$ signifies the scattering intensity of $f_n$.

The areas not illuminated by radar signals form the shadow region, denoted as \(A_{sha}\). The upper boundary of \(A_{sha}\) corresponds to the lower boundary of the aforementioned target area \(A_{tar}\), while the lower boundary of \(A_{sha}\) is determined by the mapping of the boundary beam emitted from the projection plane onto the slant range.

% This field has significant research efforts, with only a limited number of instances being presented here.
\begin{figure*}[t]
	\centering
	\includegraphics[width=1.3\columnwidth]{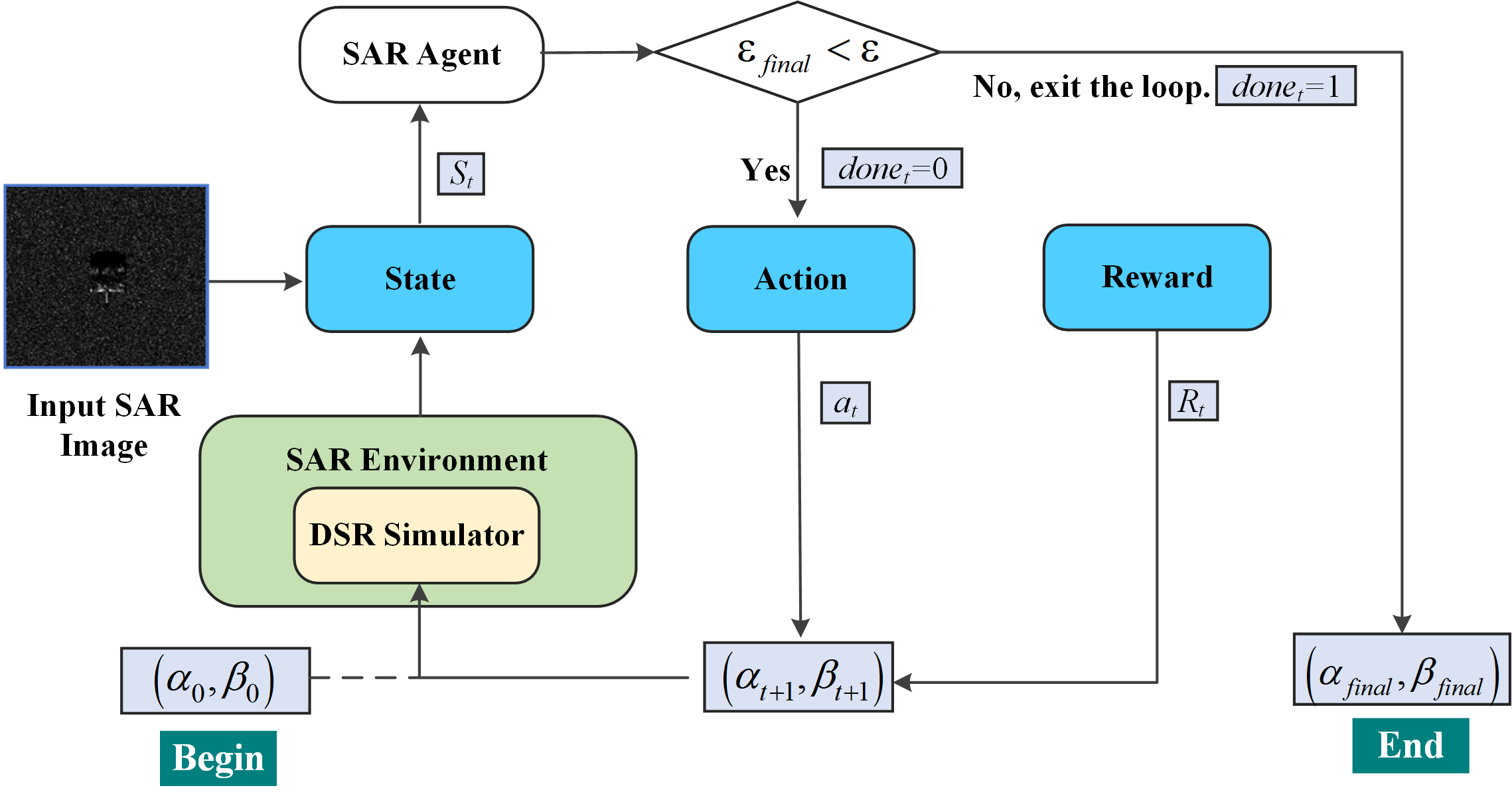}
		\caption{The workflow of our proposed view angle inversion algorithm. The DSR embedded SAR environment generates a simulated image using a given view angle $(\alpha_t, \beta_t)$ at time $t$. A state representation $s_t$ is constructed by the state construction module based on input SAR image and the simulated image. Then, upon receiving the state $s_t$, the SAR agent outputs two incremental angles $(\Delta\alpha_t, \Delta\beta_t)$ as an action, which are used to to update the currently predicted view angles to $(\alpha_t+\Delta\alpha_t, \beta_t+\Delta\beta_t)$ to form $(\alpha_{t+1}, \beta_{t+1})$.
Meanwhile, the reward construction module returns $R_t$ to the agent in proportion to the accuracy of the inversion after the action is executed.
The newly generated angles $(\alpha_{t+1}, \beta_{t+1})$ are again entered into the SAR environment for the next round of cycles.
  }
\label{RL_Model}
\end{figure*}

\begin{figure*}[t]%
\centering
\includegraphics[width=1.65\columnwidth]{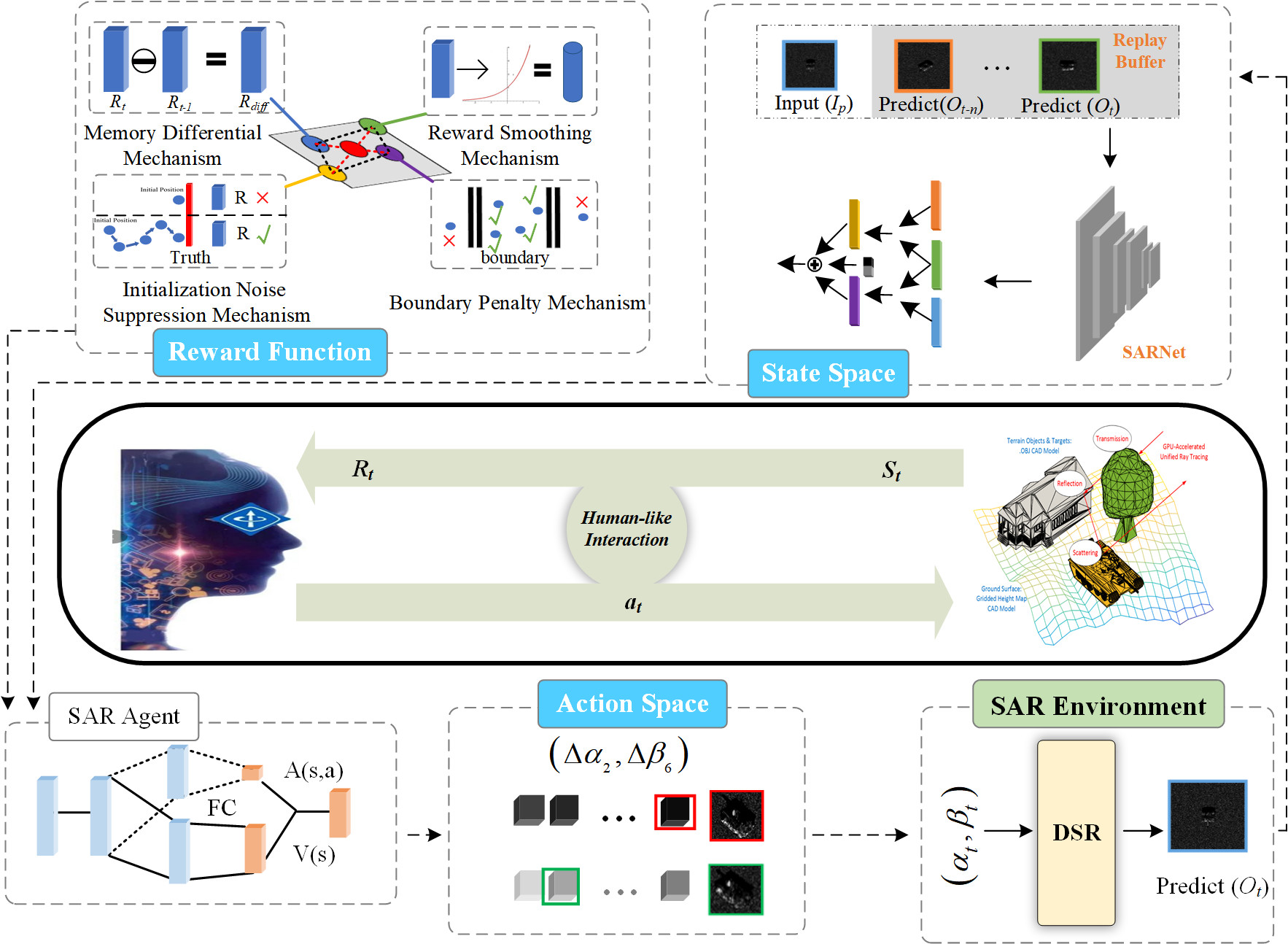}
\caption{The overall architecture of our proposed framework and the details of the different DRL modules including the SAR environment and agent, action space, state space and reward function.
%The state $s_t$ will be changed into $s_{t+1}$ after executing the action $a_t$. 
The reward $R_t$ is obtained to evaluate the inversion effect at time $t$. The option chosen by the agent depends on the currently learned policy.}
\label{overall}
\end{figure*}

\section{Methodology}
In this section, we first give an overview of the whole framework. Then the SAR environment and agent are described in detail, followed by an introduction of the DRL interactive components, including the action space, state space and reward function. Finally, we represent the training scheme of the DRL algorithm.

\subsection{Overview}
%In this paper, we propose an interactive DRL framework to reverse the radar view angles of SAR images, which facilitates the interaction between an agent and the environment with an embedded SAR simulator DSR. 

% Fig. \ref{overall} illustrates the overall architecture of our proposed method.
Fig. \ref{RL_Model} depicts the workflow of the proposed inversion algorithm. 
When given an input SAR image, incident angle and azimuth angle $(\alpha_t, \beta_t)$ at time $t$, the view angle inversion begins by generating a simulated image rendered by a DSR simulator in the SAR environment. The state construction module creates the state representation $s_t$ based on the input SAR image and the rendered image. Then, upon receiving the state $s_t$, the SAR agent outputs two incremental angles $(\Delta\alpha_t, \Delta\beta_t)$ as an action, which are used to to update the currently predicted view angles to $(\alpha_t+\Delta\alpha_t, \beta_t+\Delta\beta_t)$ to form $(\alpha_{t+1}, \beta_{t+1})$.
Meanwhile, the reward construction module returns $R_t$ to the agent in proportion to the accuracy of the inversion after the action is executed. The newly generated angles of incidence and azimuth are again entered into the SAR environment for the next round of cycles. The initial incident angle and azimuth angle $(\alpha_0, \beta_0)$ at time $t=0$ are random sampled from predefined distributions.
Following the above process, the inversion task continues until the predefined termination condition is met. 
%Interactive data is collected and stored in a replay buffer, with the agent periodically sampling and updating. 

%We present the overall architecture of the proposed framework, and the details of the different DRL modules including the SAR environment and agent, action space, state space and reward function. The SAR environment is constructed by embedding the SAR simulator DSR to generate SAR images in the intermediate process, which promotes the human-like angle prediction process. The SAR agent is designed following the architecture of Dueling DQN to output incremental angles based on the action space. The action module is designed with six discrete adjustments for each angle to quantitatively modify predicting view angles. The state module generates the state by leveraging the differences in sequential and semantic features among angle-corresponding images through the feature extraction network SARNet, which suppresses the interference of complex backgrounds and enhances sensitivity to predictive temporal changes and local information. The reward module consists of mechanisms including reward memory difference, reward smoothing, initialization noise suppression and boundary penalties, enhancing the stability and convergence of the inversion process. More details about the design of the DRL modules are provided in the following.

We delineate in Fig. \ref{overall} the comprehensive architecture of the proposed framework, encompassing various DRL modules such as the SAR environment and agent, action space, state space, and reward function. The SAR environment integrates a SAR simulator, denoted as DSR, to facilitate the generation of SAR images, thereby aiding in the emulation of human-like angle prediction. The agent is crafted using the Dueling DQN architecture, enabling it to yield incremental adjustments to the angles in the defined action space. 

The action module is designed with six discrete adjustments for each angle to quantitatively modify predicting view angles.
The state module employs SARNet, a feature extraction network, to construct the state by capitalizing on the differences in sequential and semantic features across angle-corresponding images. This approach is instrumental in diminishing the impact of complex backgrounds and amplifying sensitivity to temporal changes and local nuances. The reward module is an amalgamation of various mechanisms: reward memory difference, reward smoothing, initialization noise suppression, and boundary penalties. This confluence of strategies is aimed at bolstering the stability and expediting the convergence of the inversion process.

Subsequent sections will elaborate on the specific design elements and functional intricacies of these DRL modules.

%\subsection{Human-like Interaction between the Environment and Agent}
\subsection{SAR Environment and Agent}

\subsubsection{DSR Embedded SAR Environment}

\begin{figure*}[t]
	\centering
	\includegraphics[width=0.9\linewidth]{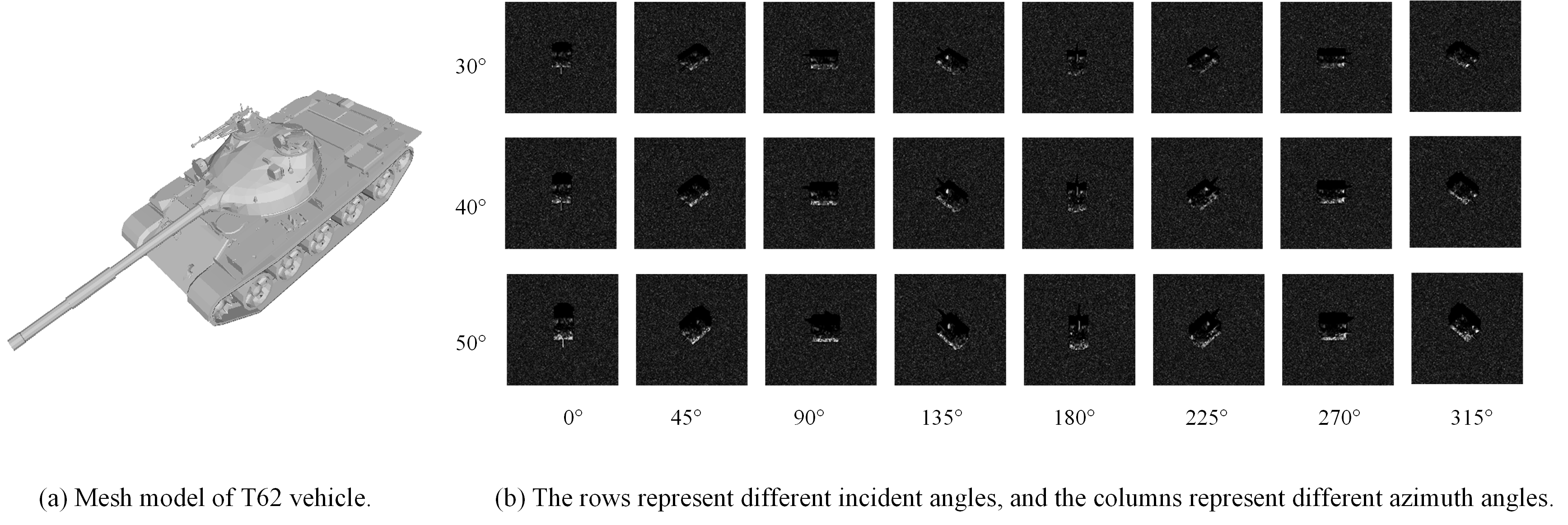}
		\caption{DSR-rendered simulated SAR images with textured T62 model at different angles.}
\label{Rendered_SAR}\end{figure*}

The method proposed in this paper embeds DSR in the interactive environment of the agent, facilitating iterative learning for the agent.  By altering the observation perspective, different SAR images can be generated by DSR.
When interacting with the agent, DSR renders arbitrary view target SAR images in real-time to promote the human-like angle prediction process, improving the agent's interactive perception and understanding of the target observation Angle.
% By altering the observation perspective, different SAR images can be generated by DSR. However, learning the angle-related imaging mechanisms from the discontinuous changes in SAR images caused by changed perspectives is a highly challenging task.
Fig. \ref{Rendered_SAR} (a) shows the 3D shape of the T62. And Fig. \ref{Rendered_SAR} (b) shows the DSR-rendered SAR images with a textured T62 model. The rows represent different incident angles, and the columns represent different azimuth angles.

\subsubsection{SAR Agent}
The SAR agent is composed of two networks: the action value network and the target action value network. These networks fulfill the role of estimating Q-values associated with each action within a given state and calculating the target Q-values during the learning process to subsequently update the parameters of the action value network.
Both networks adhere to the architecture in the Rainbow algorithm \cite{hessel2018rainbow}. 
The network ultimately produces a probability distribution encompassing various potential actions, contributing to the training trajectory of the agent.

\subsection{DRL Interactive Components}
We formulate the inverse problem of radar perspectives as a Markov decision process to discover the optimum strategy by maximizing the cumulative rewards obtained during interactive episodes. 
% Fig.\ref{overall} illustrates the overall architecture of our proposed method. 
The details of the Markov decision process are designed as follows:
\begin{itemize}
\item Action space: \textit{A} is the set of actions, representing the incremental angles $a_t$ for current view angles at time \textit{t}.

\item State space: \textit{S} is the set of states, representing the state $s_t$ generated at time \textit{t} from the observation $I_t$.

\item Reward function: $R$ is the reward function, representing the reward $R_t$ obtained by executing action $a_t$ in $s_t$ at time \textit{t}.
\end{itemize}
\subsubsection{Action Space}
%The ${a_t}$ at time \textit{t} is the step $( {\Delta {\alpha _t},\Delta {\beta _t}})$  applied to modify current view angles $(\alpha_t, \beta_t)$  quantitatively. Specifically, actions are organized in three sub-sets, symbolizing three degrees of discrete adjustment to operate view angles during interactions with the environment. For each degree of discrete adjustment, since the range of $\alpha$ and $\beta$ is different, $\alpha_i$ and $\beta_i$ are used to control the movement of angles respectively. This design allows the agent to learn both the exploration strategy with large movement in the early stage and the exploitation strategy with fine movement in the late stage, which is demonstrated to be useful for accelerating the inversion speed. In addition, a terminal action (0, 0) is devised to conclude the inverse process, ensuring the stability of the policy. 
At time $t$, the action $a_t$ is defined as the step $(\Delta \alpha_t, \Delta \beta_t)$ to adjust the current view angles $(\alpha_t, \beta_t)$ quantitatively. Actions are categorized into three subsets, representing different degrees of discrete adjustments for manipulating view angles during interaction with the environment. For each level of adjustment, given the distinct ranges of $\alpha$ and $\beta$, separate controls $\alpha_i$ and $\beta_i$ are employed for angle movement. This arrangement enables the agent to adopt an exploratory strategy with larger movements initially and a refined exploitation strategy later, enhancing the speed of inversion. Additionally, a terminal action (0, 0) is introduced to terminate the inversion process and ensure policy stability.
Then $a_t$ is as follows:
\begin{align}
{a_t} = \left\{ {(x,y)|x \in \{  \pm {\alpha _i},0\} ,y \in \{  \pm {\beta _i},0\} } \right\},{\rm{ }}i \in \left[ {1,2,3} \right],
\label{5}
\end{align}
where $\alpha_i$, $\beta_i$ are hyper-parameters. 
%In total, there are six parameters, which lead to an action space of 25. 
The hyper-parameters selected are relevant to the maximum steps in each episode, 
% are shown as follows: 
% $\left| {{\alpha _1}} \right| = 10,$ $\left| {{\alpha _2}} \right| = 2.5,$ $\left| {{\alpha _3}} \right| = 0.5,$ $\left| {{\beta _1}} \right| = 50,$ $\left| {{\beta _2}} \right| = 10,$ $\left| {{\beta _3}} \right| = 2.5$,
which enables the agent to search and make decisions in the solution space effectively and gives a good trade-off between inversion accuracy and speed.

\subsubsection{State Space}
% \begin{align}
% \left\{ {\begin{array}{*{20}{c}}
% {{\alpha _{t + 1}} = {\alpha _t} + \Delta {\alpha _t}}\\
% {{\beta _{t + 1}} = {\beta _t} + \Delta {\beta _t}}
% \label{eqr1}
% \end{array}} \right.
% \end{align}

% By modeling the above Markov decision process, the objective of radar perspectives inversion can be achieved.
%\subsubsection{SARNet-based Feature Extraction}

%\textbf{Feature Extraction Network}
% Due to the discreteness, variability and interference of target scattering properties in SAR images \cite{guo2022learning}, it is challenging for the agent to accurately extract crucial discriminative scattering features from SAR images. To overcome this issue, 

%We utilize a feature extraction network as a global discriminative feature extractor called SARNet to capture discriminative high-level semantic and low-level spatial information in SAR images, which promotes precise view angle reasoning. It is designed based on the ResNet network structure and is constructed by integrating basic network modules consisting of convolutional layers, residual connections, batch normalization, and non-linear activation functions. All the basic modules foster an incremental feature learning process. Through the assembly of multiple fundamental modules, the depth and complexity of SARNet are amplified, consequently bolstering the network's feature extraction and classification performance.

We employ SARNet, a feature extraction network, as a global discriminative feature extractor to capture high-level semantic and low-level spatial information in SAR images, thereby facilitating precise view angle reasoning. SARNet, structured on the ResNet framework, integrates convolutional layers, residual connections, batch normalization, and non-linear activation functions as its foundational modules. These modules collectively enhance incremental feature learning. As a result, the compounded depth and complexity of SARNet significantly improve its capabilities in feature extraction and classification.

%\subsubsubsection{State Construction}
%The design of the state space is essential which can provide the agent with abundant information about temporal changes, effectively enhancing the robustness of the algorithm. We first describe the observation at time step $t$ as a tuple ${o_t} = \left( {{I_p},S({\alpha _t},{\beta _t}),\left( {{\alpha _t},{\beta _t}} \right)} \right)$, where $I_p$ is the input image, $(\alpha_t,\beta_t)$ represents the current predicted view angles and $S(\alpha_t,\beta_t)$ corresponds to the rendered image associated with angle $(\alpha_t,\beta_t)$. Based on $o_t$, we construct state space by leveraging the differential of sequential and semantic features among rendered images according to adjacent view angles. Specifically, the first part of the state space is made up of the feature differential of $S(\alpha_{t-1},\beta_{t-1})$ and $S(\alpha_{t},\beta_{t})$ to provide abundant temporal variation information to enhance temporal sensitivity. The second part is the feature differential of $S(\alpha_{t},\beta_{t})$ and $I_p$ to capture semantic information about local detail changes and trajectory variations in targets caused by predicted view angles’ changes. Since the goal of the agent is to estimate the view angles of the target in SAR image, the currently estimated view angles $(\alpha_{t},\beta_{t})$ should also be included as the third part to provide direct instruction for the policy updating. 
The construction of the state space is crucial, providing the agent with comprehensive temporal information and enhancing algorithmic robustness. We define the observation at time $t$ as a tuple ${o_t} = (I_p, S(\alpha_t, \beta_t), (\alpha_t, \beta_t))$, where $I_p$ denotes the input image, $(\alpha_t, \beta_t)$ the current predicted view angles, and $S(\alpha_t, \beta_t)$ the rendered image for angles $(\alpha_t, \beta_t)$. The state space is constructed from the differential features of sequential and semantic attributes among consecutive rendered images. Specifically, it comprises the feature differential between $S(\alpha_{t-1}, \beta_{t-1})$ and $S(\alpha_{t}, \beta_{t})$ for temporal variation, and between $S(\alpha_{t}, \beta_{t})$ and $I_p$ to discern semantic changes and trajectory shifts due to variations in predicted view angles. Inclusion of the current estimated view angles $(\alpha_{t},\beta_{t})$ as the final part offers direct guidance for policy updates, aligning with the agent's objective of accurately estimating target view angles in the SAR image.
The state space is represented in the following way:
\begin{align}
{s_t} = \left( \begin{array}{l}
F\left( {S\left( {{\alpha _t},{\beta _t}} \right)} \right) - F\left( {S\left( {{\alpha _{t - 1}},{\beta _{t - 1}}} \right)} \right),\\
F\left( {S\left( {{\alpha _t},{\beta _t}} \right)} \right) - F\left( {{I_p}} \right),\left( {{\alpha _t},{\beta _t}} \right)
\end{array} \right),
\label{4}
\end{align}
where \textit{S} denotes electromagnetic simulator, \textit{F} denotes pre-trained feature extraction network. By learning the prior knowledge contained in the training SAR images, SARNet captures the robust scattering and morphological features of the target under different view angles from input SAR images.

\subsubsection{Reward Function}
%The reward function $R_t$ plays a crucial role in our inversion task, as it allows the agent to gradually optimize policy based on $s_t$ and take action. It is composed of four components: $R_{base}^t, R_1^t, R_2^t, R_3^t$. The following sections will sequentially introduce the design principles of these four parts of $R^t$.

%We first design the reward $r_1^t$ of each angle $\theta$ utilizing the difference between the inversion result and the true angle. By calculating the difference, the agent is encouraged to make movements towards the correct direction and is accordingly rewarded. When the agent's inversion result is closer to the true angle, a small negative reward is given. Instead, when the inversion result deviates further from the true angle, a large negative reward is assigned.

The reward function $R_t$ is pivotal in our inversion task, enabling the agent to iteratively refine its policy based on state $s_t$ and corresponding actions. It comprises four components: $R_{\text{base}}^t, R_1^t, R_2^t, R_3^t$. Subsequent sections delineate the design principles of these components.

Initially, we construct the reward $r_1^t$ for each angle $\theta$ by quantifying the discrepancy between the inversion outcome and the actual angle. This metric motivates the agent to adjust its actions toward the accurate angle, granting a minor negative reward for proximity and a more substantial negative reward as the deviation increases

\begin{align}
r_1^t\left( \theta  \right) =  - {L_1}\left( {{\theta _t},{\theta _r}} \right),
\label{6}
\end{align}
\begin{align}
r_{1'}^t\left( \theta  \right) =  - {L_1}\left( {{\theta _t},{\theta _0}} \right).
\label{7}
\end{align}

To expedite the convergence of the inverse policy, we have formulated $R_{\text{base}}^t$ as shown in \eqref{8}. $R_{\text{base}}^t$ employs a reward memory differential approach to encapsulate the sequential disparity between $r_1^t$ and $r_1^{t-1}$, directly steering the agent's parameter optimization. Specifically, a decrease in $R_{\text{base}}^t$ indicates an action leading away from the objective, suggesting the agent is diverging from the optimal path. Conversely, a positive $R_{\text{base}}^t$ signals alignment with the correct inversion trajectory, prompting continued exploration in that vector
\begin{align}
R_{\text{base}}^t = r_1^t\left( \theta  \right) - r_1^{t-1}\left( \theta  \right).
\label{8}
\end{align}

To mitigate the inversion process's instability, potentially causing erratic policy performance, we introduce a negative exponential reward $r_2$. This strategy ensures smooth, gradual adjustments towards the target value and discourages abrupt, unstable actions. Moreover, to avert the agent's fixation on optimizing a solitary parameter, we have structured the reward function to foster simultaneous optimization of multiple parameters, specifically $\alpha$ and $\beta$, with distinct parameters ${\eta_1} =  - 0.05$ and ${\eta_2} =  - 0.2$ respectively

\begin{align}
r_2(o|\eta) = \exp(\eta \cdot o),
\label{9}
\end{align}
\begin{align}
R_1^t = r_2(r_1^t(\theta)|\eta).
\label{R_1}
\end{align}

In cases where initial view angles are suboptimal, the replay mechanism's efficiency diminishes, leading to reward fluctuations within an episode. Notably, when starting near the true value, relying solely on $R_1^t$ diminishes exploration, risking convergence to local optima. To counteract the variability from initial conditions, an auxiliary reward, $R_2^t$, is introduced as indicated in \eqref{11}.
\begin{align}
{R_2^t} = 5\sum\limits_{i = 1}^3 {A\left( {\alpha |{\rho _i},{\varphi _i}} \right)}  \cdot A\left( {\beta |{\nu _i},{\omega _i}} \right),
\label{11}
\end{align}
\[
{s.t.\,\,{\rm{ }}A\left( {\theta |{\kappa_i},{\varsigma_i}} \right) = \left\{ {\begin{array}{*{20}{l}}
{\begin{array}{*{20}{l}}
1&\rm{if}\quad{\left| {r_1^t\left( \theta  \right)} \right| < {\rho _i}\,\& \,\left| {r_{1'}^t\left( \theta  \right)} \right| > {\varphi _i}},
\end{array}}\\
{\begin{array}{*{20}{l}}
0&{{\rm{otherwise}}}
\end{array}},
\end{array}} \right.}
\]
where ${\rho _i},{\varphi _i},{\nu _i},{\omega _i}$ are hyper-parameters.
Here, $A(\theta |{\kappa_i},{\varsigma_i})$ activates only when precise conditions of proximity and improvement are met, encouraging comprehensive optimization and escalating rewards as the challenge of adjustment intensifies. This mechanism averts scenarios of substantial rewards for minimal or no actual improvement.

To deter the agent from straying beyond the beneficial parameter space, thereby disrupting the exploration continuum, we impose a penalty through $R_3^t$. Set to $-10$, this punitive measure bounds the action outputs, fostering a focused, effective exploration within the permissible parameter range
\begin{align}
R_t = R_{\text{base}}^t + R_1^t + R_2^t + R_3^t.
\label{12}
\end{align}

The intricate composition of $R_t$ effectively amalgamates individual rewards, counteracting policy non-convergence issues common in sparse reward scenarios in RL. It ensures methodical, quality actions at each iteration, enhancing the learning trajectory and propelling the agent towards peak performance.

\subsection{Training Scheme of the DRL Algorithm}
%\subsubsection{DRL Algorithm Selection of SAR Agent}

%The objective of our task is to deduce the radar view angles $[\alpha, \beta]$ inherent to the SAR image \textit{X}, which is achieved by employing DRL to learn the mapping policy ${F_\Theta }=X\to[\alpha, \beta ]$. In consideration of the discrete action space, we adopt the Rainbow algorithm \cite{hessel2018rainbow}, which aligns well with discrete action spaces \cite{mnih2013playing} and intensifies the agent’s learning efficacy and bolsters training stability. 

Our objective is to infer the radar view angles $[\alpha, \beta]$ inherent to the SAR image $X$, utilizing Deep Reinforcement Learning (DRL) to learn the mapping policy $F_\Theta: X \to [\alpha, \beta]$. Given the discrete nature of the action space, the Rainbow algorithm \cite{hessel2018rainbow} is employed for its compatibility with discrete actions \cite{mnih2013playing} and its enhancement of learning efficiency and training stability.

We initialize the action value network $Q(s, a|\theta )$ parameters as $\theta$ and the target action value network $Q'(s, a|\theta')$ parameters as $\theta'$. The agent then proceeds with iterative interactions across multiple episodes. In each episode, the Q-network receives state $s_t$ and outputs probabilities for various actions. It follows an $\epsilon$-greedy strategy, exploring with probability $\epsilon$ and exploiting with $1-\epsilon$. Over time, this strategy is refined by linearly decreasing $\epsilon$, incrementally favoring exploitation. Interactions terminate upon reaching the environment's maximum step count, signaling a terminal state. Each episode's trajectories are stored in an experience replay buffer, adopting a prioritized strategy from the Rainbow algorithm for efficient learning by extracting high-priority samples based on their estimated value. The parameter update for the Q-network, influenced by a discount factor $\gamma$, is formalized in \eqref{13}
\begin{align}
loss = {\left( \begin{array}{l}
{R_t} + \gamma {Q_{\theta '}}\left( {{s_{t + 1}},\mathop {{\mathop{\rm argmax}\nolimits} }\limits_{{a^\prime }} {Q_\theta }\left( {{s_{t + 1}},{a^\prime }} \right)} \right) - \\
{Q_\theta }\left( {{s_t},{a_t}} \right)
\end{array} \right)^2}.
\label{13}
\end{align}

Throughout the testing phase, the refined network parameters are directly employed to yield precise probability estimates for all possible action executions corresponding to a provided state. After several iterations, the final inversion angles can be gained.

\section{Experiments}
In this section, we introduce the experimental setup and the implementation details of agent training and testing. Furthermore, we undertake a thorough assessment of our proposed algorithm, encompassing extensive quantitative experiments, evaluation of action strategy, ablation analyses and domain adaptation experiments.
% \begin{figure}[t]
% \vspace{-0.4cm}
% 	\centering
% 		\includegraphics[width = 0.85\columnwidth]{Image/T62_3.png}
% 		\caption{Mesh model of T62 vehicle.}
% \label{T62}\end{figure}
\subsection{Experimental Dataset} 
\subsubsection{Simulated Dataset} 
%The simulated SAR images employed during the training and testing phases of this study are synthesized via DSR rendering. Following the definition of the incident angle in this paper, once the incident angle exceeds a predefined threshold, the projected position of the target will extend beyond the region of the image, resulting in the loss of pertinent information concerning the target’s configuration and contour within the projected depiction, which in return affects the agent to acquire a coherent inversion strategy. To uphold the fidelity of the rendered images, the agent's angles are defined as combinations of arbitrary angles residing within the spectrum of incidence angles [30\degree, 75\degree] and azimuth angles [0\degree, 359\degree]. The dimensions of the rendered images are standardized to 128×128 pixels. 
In this study, simulated SAR images for training and testing are synthesized using DSR rendering. According to the incident angle criteria defined herein, surpassing a certain threshold causes the target's projected position to extend outside the image bounds, leading to loss of critical target configuration and contour data. This impacts the agent's ability to develop a consistent inversion strategy. To maintain image fidelity, the agent's view angles are set within a range of incidence angles [30\degree, 75\degree] and azimuth angles [0\degree, 359\degree]. All rendered images are standardized to a resolution of $128\times 128$ pixels.

\subsubsection{Real Dataset} 
%The MSTAR dataset is selected as the real dataset to validate the generalization ability of the proposed algorithm \cite{MSTAR}. It comprises different vehicle types at various viewing angles. The imaging mode is a spotlight with a resolution of 0.3 × 0.3m, utilizing a 128 × 128 pixel chip. The azimuth angles range from 0 to 360°, with intervals approximately 1-2°. The depression angle is 17° and 15°.
The MSTAR dataset is utilized to validate the generalization ability of the proposed algorithm \cite{MSTAR}. It encompasses various vehicle types imaged at multiple viewing angles. The images are captured in spotlight mode with a resolution of 0.3m x 0.3m and formatted into 128 x 128 pixel chips. Azimuth angles span from 0 to 360° at approximately 1-2° intervals. The dataset includes depression angles of 17° and 15°.

\subsection{Experiment Setup}

\subsubsection{Experiment Model} This study selects T62 as a 3D model and employs DSR rendering this shape to validate the efficacy of our algorithm. 

\begin{figure}[t]
	\centering
		\includegraphics[width = 0.85\columnwidth]{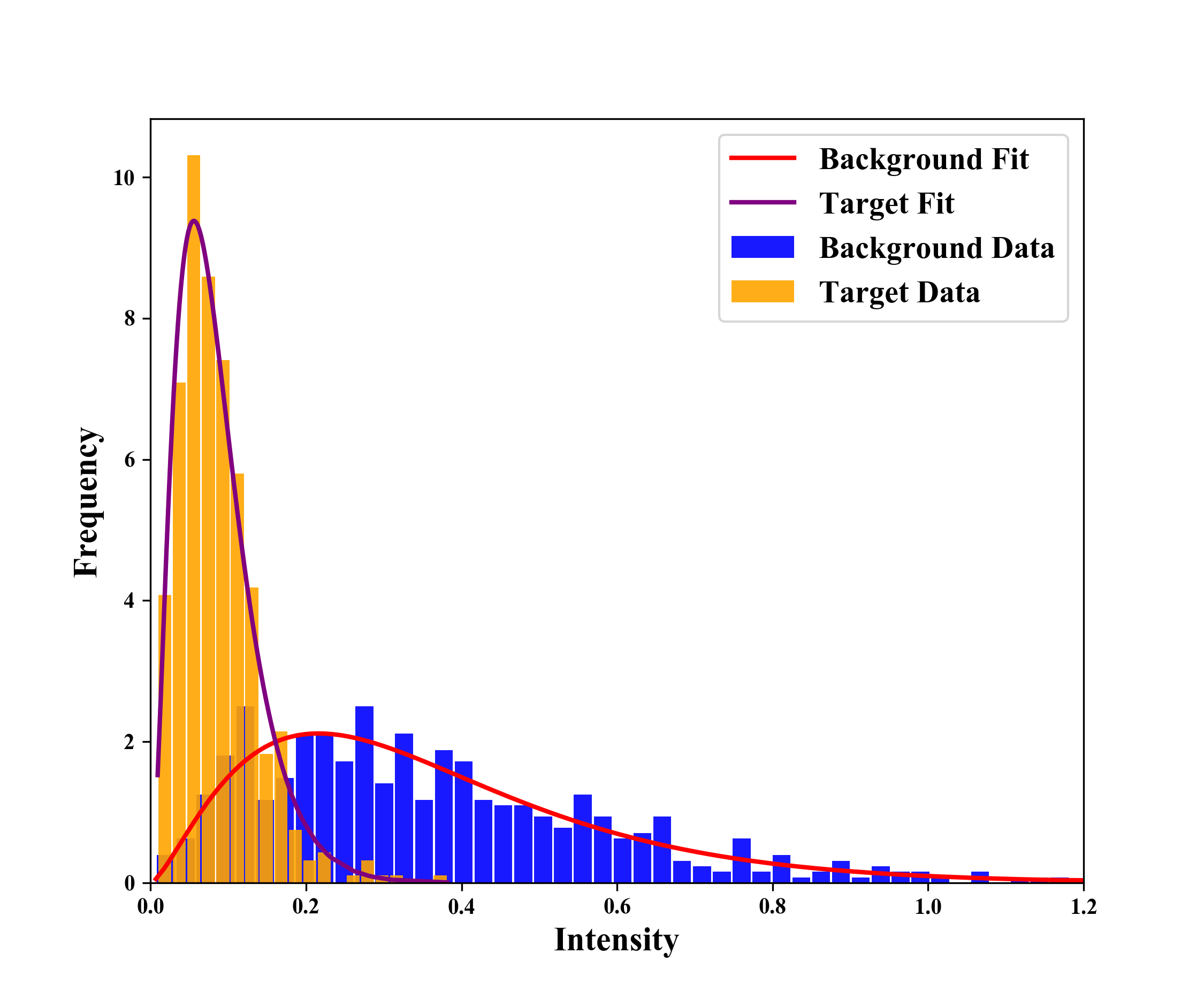}
		\caption{Gamma Distribution Fitting.}
\label{figure-9}\end{figure}

Due to varying scattering characteristics of objects and observation conditions, the imaging results for the same target also differ. To generate realistic simulated SAR images, the standard gamma distributions are used to extract scattering textures of objects in the MSTAR dataset \cite{MSTAR}. Firstly, the SAR image is segmented into the target region and background region with the segment anything algorithm \cite{kirillov2023segment}. Next, gamma distributions are independently fitted to the texture attributes of these two distinct regions (see Fig. \ref{figure-9}), and the parameters of the two distributions are tabulated within Table \ref{TABLE1}. The values obtained through stochastic sampling from the above distributions are then designated as textures attributed to the surface of the 3D target and the ground and hence a 3D model enriched with scattering textures is obtained.

% \begin{table}[t]
% \renewcommand{\arraystretch}{1.2}
% \setlength{\tabcolsep}{18pt}
% \caption{DRL Parameters}
% \centering
% \begin{adjustbox}{center}
% \begin{tabular}{c|c}
% \hline
% \makebox[0.07\textwidth][c]{\textbf{RL Parameters}} & \makebox[0.04\textwidth][c]{\textbf{Value}} \\
% \hline
% Buffer size & 50,000 \\
% Batch size & 256 \\
% Learning Rate & 0.00001 \\
% Initial $\epsilon$ & 0.5 \\
% Final $\epsilon$ & 0.01 \\
% $\gamma$ & 0.96 \\
% Update Frequency & 200 \\
% \hline
% \end{tabular}
% \end{adjustbox}
% \label{RL_parameters}
% \end{table}

\subsubsection{Experiment Configuration}
All network models discussed in this article are implemented using PyTorch, a Python-based DL framework. The training and testing of these models are conducted on an NVIDIA GeForce RTX 2080 Ti graphics card, utilizing CUDA 10.1 and cuDNN 8.0.5 for accelerated computations. During the training phase, the model parameters are optimized using the Adam optimization algorithm \cite{kingma2014adam}.
In the RL framework, the capacity of the replay buffer is 50,000, allowing the agent to store data until the buffer is full. Both agent networks are trained with a batch size of 256 and a learning rate of $1e-5$. To facilitate exploration, an $\epsilon$-greedy strategy is employed, and $\epsilon$ is linearly annealed from 0.5 to 0.01 according to this policy. The discount factor $\gamma$ is set to 0.96, and the target network is updated every 200 steps.

The feature extraction network is trained on DSR-rendered simulated SAR data 
generated by uniform sampling within pre-defined angle distribution regions, which ensures that the generated samples encompass a variety of real-world SAR scenarios. Two distributions are formulated as:
\begin{align}
D_{\alpha} \sim D_{\text {uniform }}(a, b \mid \rho), a=35, b=75, \rho=5 ,
\label{2}
\end{align}
\begin{align}
D_{\beta} \sim D_{\text {uniform }}(c, d \mid \upsilon), c=0, d=360, \upsilon=5 ,
\label{3}
\end{align}
where $a$, $b$ are the starting and ending angles of distribution $D_{\alpha}$ respectively, $\rho$ denotes the sampling interval. The parameter definition of distribution $D_{\beta}$ is as same as distribution $D_{\alpha}$. 

\subsubsection{Evaluation Metrics}
To assess the angle inversion performance of the agent, we employ kinds of metrics such as $MAE$, $MAPE$, $RMSE$ and $MedAE$. Denote by ${\tilde P^i}$ and ${P^i}$ the inversion estimated and the true angles for the $i_{th}$ sample, respectively. Denote by $n$ the total number of samples.
\begin{itemize}
\item $MAE$ assesses the disparity between predicted values and actual observations and assigns equal weight to all samples, which evaluates the fundamental performance of our model
\begin{align}
{MAE} = \frac{1}{n}\sum_{i=1}^n| {{{\tilde P}^i} - {P^i}} |.
\label{14}
\end{align}
\item $MAPE$ calculates the average percentage error of predicted values relative to the actual observations, which emphasizes the assessment of predictive performance where true values exhibit significant variations
\begin{align}
MAPE =\frac{1}{n} \sum_{i=1}^n\left|\frac{{P^i}-{{\tilde P}^i}}{P^i}\right|.
\label{Metric3}
\end{align}
\item $RMSE$ measures the average size of the agent's prediction error in the form of root-mean-square, which primarily evaluates the generalization ability of our model for outliers
\begin{align}
RMSE=\sqrt{\frac{1}{n} \sum_{i=1}^n({P^i}-{{\tilde P}^i})^2}.
\label{Metric4}
\end{align}
\item $MedAE$ quantifies the median absolute difference between actual observations and their corresponding predictions, which concentrates on the accuracy of predictions for inliers
\begin{align}
MedAE = median(|{{{\tilde P}^i} - {P^i}} |).
\label{Metric5}
\end{align}
% \item $R^2$ quantifies the effectiveness of independent variables in predicting the dependent variables, which focuses on the model's ability to explain variations in the data. 
% \begin{align}
% R^2 =1-\frac{\sum_{i=0}^m\left({P^i}-{{\tilde P}^i}\right)^2}{\sum_{i=0}^m\left({P^i}-\bar{P^i}\right)^2}
% \label{Metric4}
% \end{align}
\end{itemize}
% $MSLE$ finds application in regression tasks, measuring the mean squared error between predicted values and the natural logarithm of actual observations. 
% \begin{align}
% {AE_{angle}} = | {{{\tilde P}^i} - {P^i}} |
% \label{14}
% \end{align}

% \begin{align}
% MSLE = \frac{1}{n} \sum_{i=1}^n(\log(1+{P^i})-\log(1+{{\tilde P}^i}))^2
% \label{Metric4}
% \end{align}

\begin{table}[t]
\renewcommand{\arraystretch}{1.2}
\setlength{\tabcolsep}{6pt}
\caption{Parameters of the Gamma distributions\label{tab:table1}.}
\centering
\begin{adjustbox}{center}
\begin{tabular}{c|c|c}
\hline
\textbf{Region} & \textbf{Shape Parameters} &\textbf{Scale Parameters} \\
\hline
Target & 2.311 & 0.162\\
Background & 2.867 & 0.029\\
\hline
\end{tabular}
\end{adjustbox}
\label{TABLE1}
\end{table}

\begin{table*}[t]
\renewcommand{\arraystretch}{1.5}
\setlength{\tabcolsep}{10pt}
\caption{Metrics Comparison Results of different algorithms on DSR-rendered SAR Images\label{tab:table2}. The Best Result for Each Metric is Highlighted in Bold.}
\centering
\begin{center}
\resizebox{\linewidth}{!}{
\begin{tabular}{c|c|c|c|c|c|c|c}
\hline
Method 
& \textbf{$MAE_{\alpha}$} 
& \textbf{$MAE_{\beta}$} 
& \textbf{$MAE_{mean}$} 
& \textbf{$MAPE$} 
& \textbf{$RMSE$} 
& \textbf{$MedAE$} 
% & \textbf{$R^2$} 
& Run Time (s) \\
\hline 
GA\cite{holland1992adaptation} & 4.580 & 8.612 & 6.596 & 0.086 & 7.489 & 5.892  &  115.600\\
PSO\cite{kennedy1995particle} & 6.249 & 11.416 & 8.833 & 0.118 & 9.890 & 3.797 &  58.912\\
DL\cite{he2016deep} & 2.052 & 3.993 & 3.023 &0.040 & 3.578 &2.078  & \textbf{0.160}\\
{DRL (Ours)} & 1.594 & 3.797 & 2.696 & 0.043 & 3.168 & 2.000   & 2.282\\
{DL+DRL (Ours)} & \textbf{1.282} & \textbf{1.960} & \textbf{1.621} & \textbf{0.030} & \textbf{2.070} & \textbf{1.239}  &0.508\\
\hline
\end{tabular}}
\end{center}
\label{Ablation_1}
\end{table*}

\subsection{Inversion Strategy Evaluation}
This segment presents reward curves that illustrate the progression observed during the training phase of the proposed algorithm. In addition, the performance of the agent in the testing phase is also validated.

% \subsubsection{Testing of Agent}
\subsubsection{Comparative experiments}
In order to further verify the performance of our proposed algorithm, we compare it with the GA algorithm, PSO algorithm, DL algorithm and DL combined with the DRL algorithm. The basic idea of GA and PSO algorithms is to model the view angles inversion problem as an optimization problem. 
Its objective is to minimize the difference between the input image and the rendered image associated with inverse angles, which is equal to finding the minimum of the objective function. 
% The experiment shows that if the fitness function is directly defined as $L_1$ loss of these two images, the optimization result is poor. 
To accelerate the optimization process and enhance its precision, the fitness functions of the two algorithms are designed as the feature difference between two images as \eqref{PSO_GA}:
\begin{align}J(\alpha ,\beta ) = {\left\| {F\left( {S\left( {\alpha ,\beta } \right)} \right) - F\left( {S\left( {{\alpha _p},{\beta _p}} \right)} \right)} \right\|_1}.
\label{PSO_GA}
\end{align}

% \begin{table}[h]
% \renewcommand{\arraystretch}{1.2}
% \setlength{\tabcolsep}{18pt}
% \caption{Parameters of GA and PSO}
% \centering
% \begin{adjustbox}{center}
% \begin{tabular}{c|c|c}
% \hline
% \makebox[0.03\textwidth][c]{\textbf{Algorithm}}& 
% \makebox[0.05\textwidth][c]{\textbf{Parameters}} & 
% \makebox[0.05\textwidth][c]{\textbf{Value}} \\
% \hline
%   \multirow{4}{*}{GA} & Population &20 \\
%  & Iteration Number & 10 \\
%  & Crossover Rate  & 0.9 \\
%  & Mutation Rate & 0.1 \\
% \hline
%   \multirow{5}{*}{PSO} & Pop size &20 \\
%  & Swarm Size & 20\\
%  & Inertia Weight &0.5 \\
%  & Learning Factor1 & 2 \\
%  & Learning Factor2 & 2 \\
%  & Iteration Number & 10 \\

% \hline
% \end{tabular}
% \end{adjustbox}
% \label{RL_parameters}
% \end{table}

In contrast to the mechanism of the above optimization algorithms, the DL algorithm aims to employ neural networks to fit the inverse mapping function of our task. Then a dataset consisting of 576 DSR-rendered T62 target SAR images is created, encompassing $\alpha$ within [35\degree, 70\degree] in 5-degree intervals and $\beta$ within [0\degree, 360\degree) in 5-degree intervals. The DL network structure aligns with the previously mentioned pre-trained model network structure. 

%Some researchers have also embraced a conjoined methodology that combined DL and DRL. In this fusion, RL is utilized to meticulously refine DL models, such as ChatGPT, which operates within an RL framework through human feedback. This strategy is employed arises from the fact that DL merely trains to create a complex mapping without exploring of the inherent optimization policy. Moreover, due to the limited interpretability and terrible robustness of DL, it's challenging to incorporate physical priors into models. Meanwhile, relying solely on RL agents to explore valid solution space is time-consuming. Therefore, this paper also compares the DRL algorithm with the DL+DRL approach. Specifically, it employs the outputs of DL as initial randomized angles for DRL. By fine-tuning DL output angles, the agent's starting estimated values are close to the true solution, thereby curtailing the exploration duration of the agent. 

Several researchers have adopted an integrated methodology that merges DL and DRL. In this amalgamation, RL is applied to enhance DL models meticulously, exemplified by systems like ChatGPT, which function within an RL paradigm augmented by human feedback. This approach stems from the recognition that DL primarily focuses on generating complex mappings without sufficiently addressing the underlying optimization strategies. Furthermore, the limited interpretability and robustness of DL pose significant challenges in integrating physical priors into models. Conversely, relying exclusively on RL agents to navigate the solution space proves to be an extensive process. Consequently, this paper contrasts the DRL algorithm with the combined DL and DRL strategy. Specifically, it utilizes DL-derived outputs as the initial states for DRL, refining these outputs to align the agent's initial estimates more closely with the actual solution, thus reducing the time required for exploration.
% The hyper-parameters in action space of the DRL framework are set as follows: $\left| {{\alpha _1}} \right| = 1$, $\left| {{\alpha _2}} \right| = 0.5$, $\left| {{\alpha _3}} \right| = 0.002$, $\left| {{\beta _1}} \right| = 1$, $\left| {{\beta _2}} \right| = 0.5$, $\left| {{\beta _3}} \right| = 0.002$.

Table \ref{Ablation_1} presents a comparative analysis of the mean inversion performance, considering the accuracy and time consumption. The best result for each metric is highlighted in bold. In terms of accuracy, the DRL method outperforms both optimization algorithms and DL methods, yielding the smallest error in $MAE_\alpha$, $MAE_\beta$, $MAE$, $RMSE$ and $MedAE$, where $MAE$ represents the average inversion error of two angles. Compared to the DL method, a reduction of 0.327, 0.410 and 0.078 on $MAE$, $RMSE$ and $MedAE$ can be achieved by our DRL method. The $MAE$ and $RMSE$ of DRL are merely 2.696 and 3.168 degrees, showcasing a significant advantage on all samples and outliers. This outcome substantiates the higher precision of the DRL inversion algorithm, primarily attributable to the well-designed $s_t$ in RL that captures the physical features and semantic information of SAR images, effectively harnessing temporal information as well. Compared to the DL and DRL methods, our DL+DRL method gets state of the art in $MAE_\alpha$, $MAE_\beta$, $MAE$, $RMSE$ and $MedAE$ metrics. The results demonstrate that our DL+DRL method uses the robust feature extraction of DL and the strategic decision-making ability of RL. This synergy offers a more effective method for addressing complex, dynamic problems requiring long-term planning and strategy formulation. Regarding the processing speed of models, the combination of DL and DRL requires only 10 interaction steps to achieve a more accurate inversion result. This is due to the swift convergence achieved by replacing random initialization with DL-predicted angles, thus localizing the solution near optimality, coupled with a finer granularity in the action space settings.
\vspace{-0.1cm}
% In terms of the inversion speed, machine learning methods demonstrate a conspicuous advantage over traditional optimization algorithms. 
% This disparity mainly stems from the iterative nature of optimization algorithms, where the lengthy duration of individual iterations coupled with multiple iterations required for convergence leads to prolonged overall processing time. In contrast, machine learning network outputs are swift; DL, requiring only a single output, boasts the shortest execution time. Although DRL entails multiple iterations, it also accomplishes the inversion within a relatively short timeframe. The integrated DL-DRL approach exhibits time consumption comparable to that of the DL algorithm, while reaches the highest inversion accuracy at the same time.
% Compared to utilizing DRL alone, 

\begin{table*}[t]
\renewcommand{\arraystretch}{1.5}
\setlength{\tabcolsep}{10pt}
\caption{Metrics Comparison Results of different algorithms on MSTAR. The Best Result for Each Metric is Highlighted in Bold.}
\centering
\begin{center}
\resizebox{\linewidth}{!}{
\begin{tabular}{c|c|c|c|c|c|c|c}
\hline
Method 
& \textbf{$MAE_{\alpha}$} 
& \textbf{$MAE_{\beta}$} 
& \textbf{$MAE_{mean}$} 
& \textbf{$MAPE$} 
& \textbf{$RMSE$} 
& \textbf{$MedAE$}  
% & \textbf{$R^2$} 
& Run Time (s)\\
\hline 
GA\cite{holland1992adaptation} & 8.353 & 15.574& 11.963 & 0.109 & 13.551 & 9.016 &  94.639 \\
PSO\cite{kennedy1995particle} & 5.306 &12.396 & 8.851 & 0.086 & 10.464 &  4.399 & 106.667  \\
DL\cite{he2016deep} & 1.359 &6.859 & 4.109 & 0.037 & 5.087 & \textbf{1.499}  &  \textbf{0.157} \\
{DRL (Ours)} & 1.282 & 6.169& 3.726 &\textbf{0.033}  &4.403  & 2.895  & 1.350\\
{DL+DRL (Ours)}& \textbf{1.099} & \textbf{5.618}& \textbf{3.359} &0.034  & \textbf{4.163} & 2.479 &  1.049 \\
\hline
\end{tabular}}
\end{center}
\label{doman1}
\end{table*}

\subsubsection{Domain Adaptation}

Discrepancies between SAR simulators and practical radar sensors create significant domain gaps between simulated and real data, often yielding suboptimal model performance with real-world data. Nevertheless, the DRL interaction framework demonstrates robust generalization capabilities across domains. Notably, by fine-tuning with a minimal set of target domain samples, the agent effectively explores similarities between the training and target domains through interaction, thereby acquiring adaptive cross-domain policies. To empirically validate the generalization of our proposed framework, a subset of real T62 category images from the MSTAR dataset \cite{MSTAR} are integrated into the training phase. In each episode, a randomly selected real image is inputted, followed by iterations using DSR-simulated data. The fine-tuned agent is then employed to reconstruct the radar perspective of previously unseen real images. 

The results of these domain adaptation experiments are delineated in Table \ref{doman1}.
The table reveals that the predicted $MAE$ and $RMSE$ values are relatively low due to the MSTAR dataset's narrow range of incidence angles, while the azimuth angles span $ [  {{0}^ \circ },{{360}^ \circ }]$, augmenting prediction complexity and error margins. Despite increased deviation in measured data, the overall inversion error remains modest, underscoring the method's robustness in managing cross-domain datasets. $MedAE$, an outlier-resistant metric, shows that excluding outliers significantly enhances DL performance, thereby confirming the superior generalization of our DRL methods over conventional DL techniques in outlier scenarios.

\subsection{Evaluation of Agent's Action Strategy}

\begin{figure}[t]
	\centering
		\includegraphics[width = \linewidth]{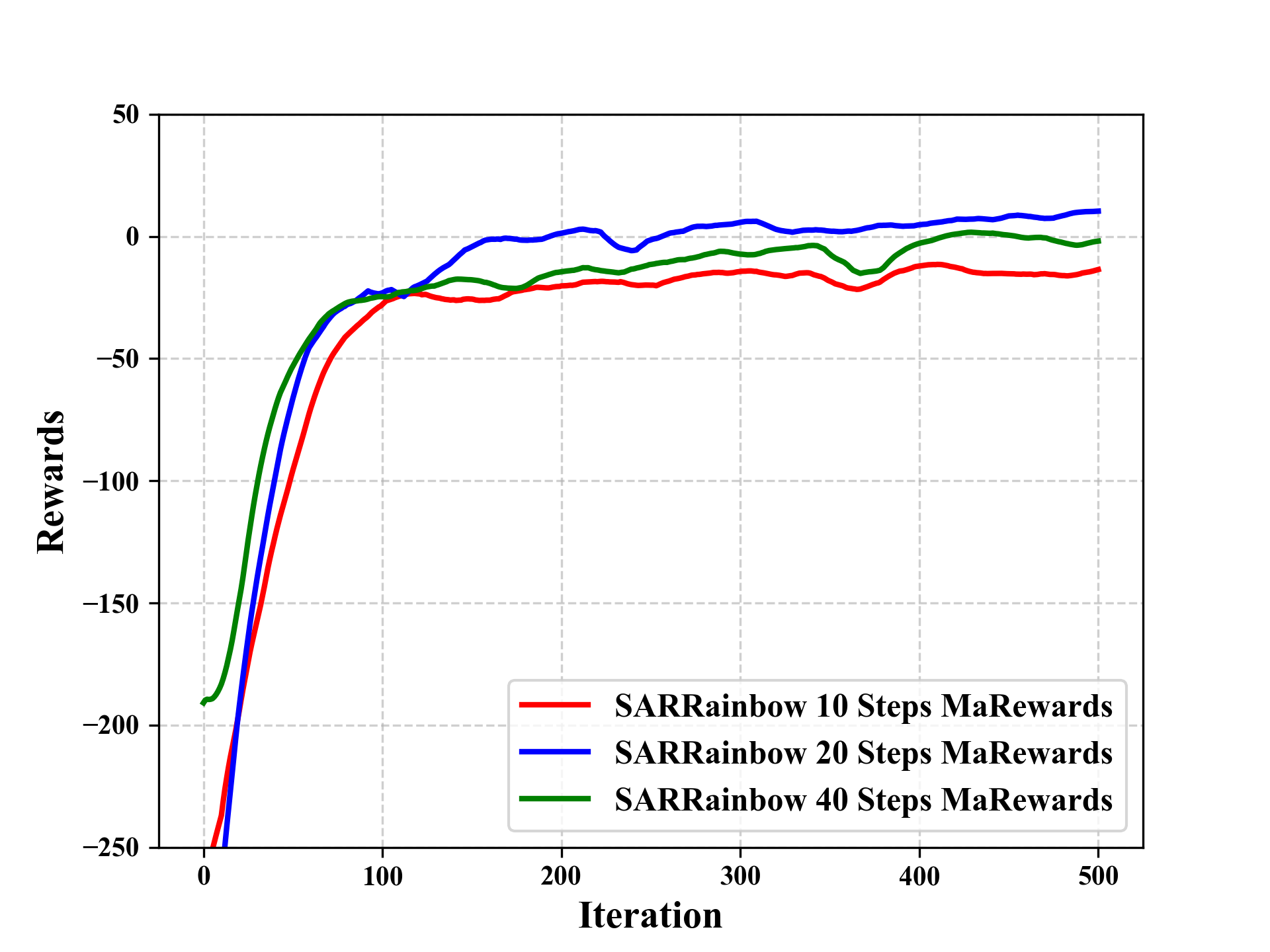}
		\caption{Reward curves of three agents with different steps in the training phase.}
\label{fig_7}\end{figure}

\subsubsection{Training Inversion Agent}
In order to ensure the agent’s convergence, the maximum training iteration is set to 500. When reaches the maximum, the agent concludes its training. Within an episode, termination is triggered under two circumstances: if the interaction steps attain the pre-defined maximum step in each episode, or if $MAE_{angle}$ descends below 2 degrees. The interaction steps are assigned distinct values in each episode for different training times, specifically 10, 20 and 40 steps respectively. 

The reward curves corresponding to these three agents of different interaction steps are visually presented in Fig. \ref{fig_7}.
In Fig. \ref{7}, all three curves adhere to the expected trajectory: as training progresses, the cumulative reward for each interaction quickly rises and stabilizes around the 120\emph{th} iteration. In subsequent iterations, some minimal fluctuations exist attributable to the inherent randomness in the selection of initial angles. Despite these minute fluctuations, the reward curves remain relatively stable. This pattern confirms the effectiveness and robustness of our task modeling.

\subsubsection{Visualization of MAE}
\begin{figure}[t]
	\centering
		\includegraphics[width =  \linewidth]{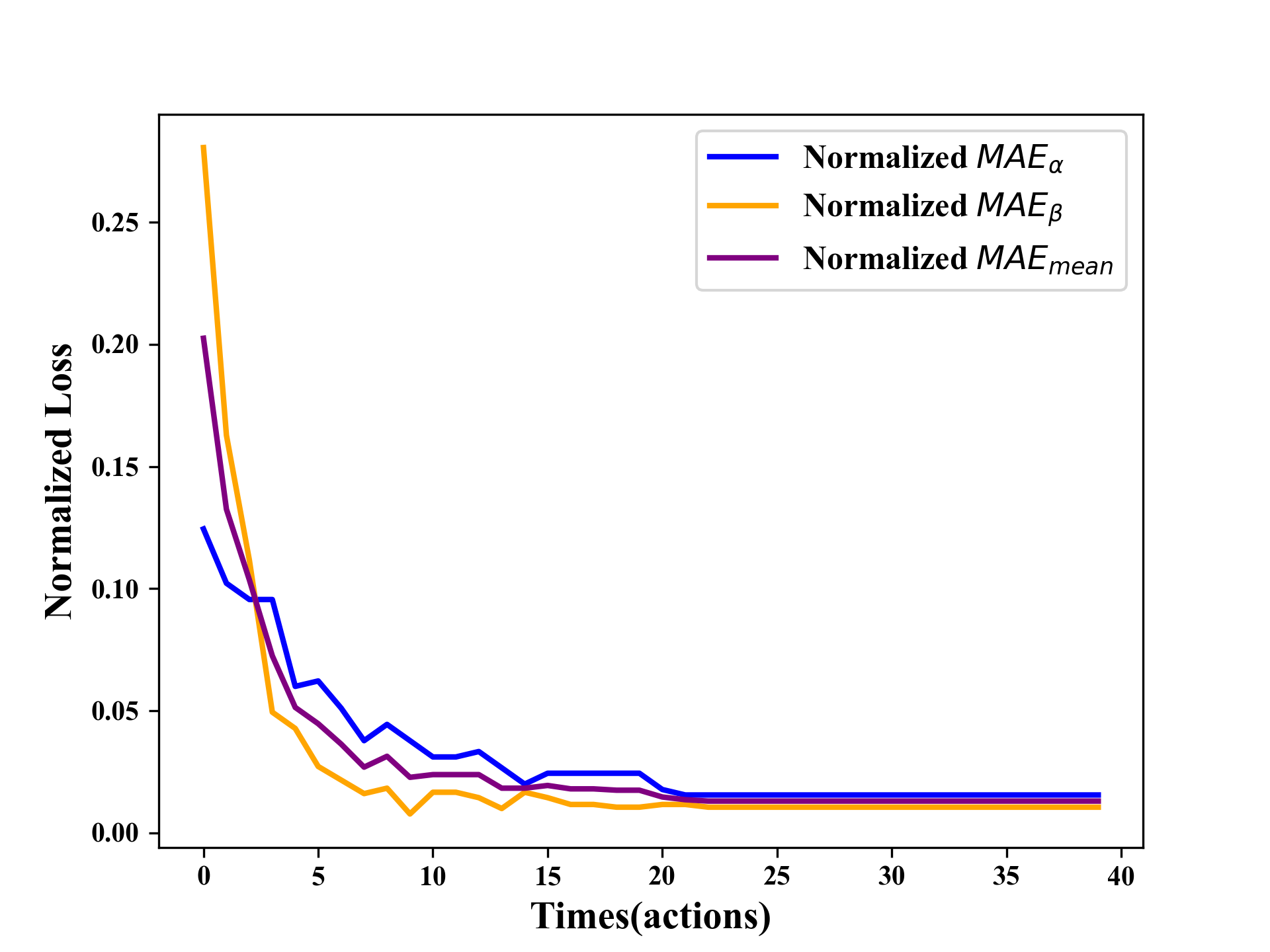}
		\caption{Average normalized change in $MAE_{angle}$ with increasing interaction steps during a single episode.}
\label{10}\end{figure}

We visualized the normalized average change in $MAE_{angle}$ with increasing interaction steps during a single episode. This metric intuitively reflects the normalized discrepancy between inversion results and ground truth. As depicted in Fig. \ref{10}, as the number of interaction steps increases, the inversion results gradually converge towards the ground truth. When the average action selection number reaches around 5, the \textit{MAE} converges to a relatively low value. Upon further increasing the interaction steps to approximately 15, the agent nearly achieves complete convergence.

\subsubsection{Visualization of Agent's Action Selection}
According to the design of the action space in this paper, it is evident that three sets of action adjustment parameters are established, representing three adjustment magnitudes for two angles. To investigate whether the agent has learned a phased exploration strategy, the iterative process of each episode is evenly divided into the early stage and the late stage based on the total number of steps taken by the agent. Theoretically, in the early stage, the agent tends to adopt a larger action adjustment magnitude to narrow down the solution space swiftly. In the late stage, the agent employs smaller actions for fine-tuning to enhance inversion accuracy. 

\begin{figure}[th]
	\centering
		\includegraphics[width = \linewidth]{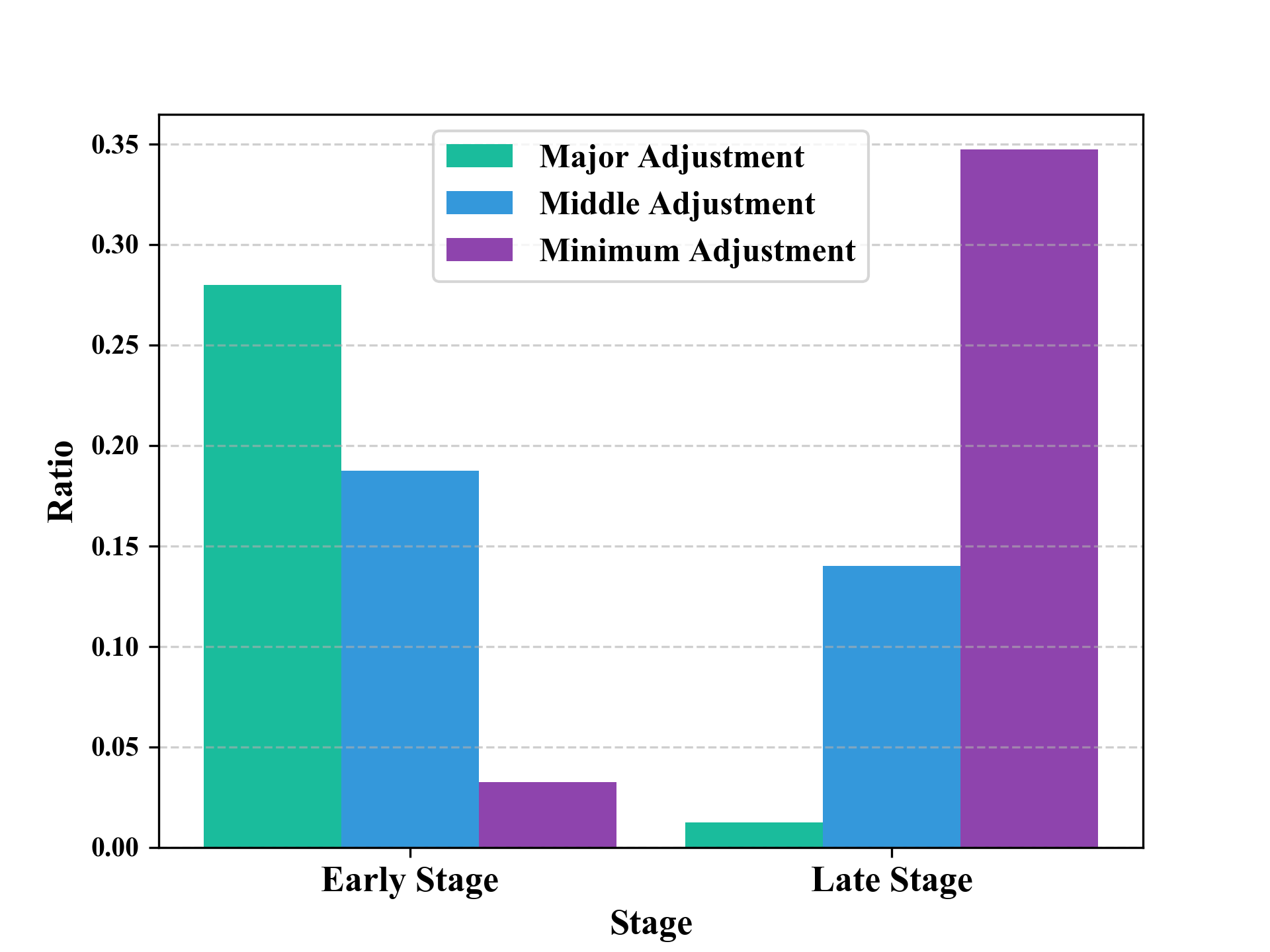}
		\caption{Frequency for different adjustments of the actions generated by the SAR agent.}
\label{fig_9}\end{figure}

Fig. \ref{fig_9} visualizes this process, demonstrating that the proportion of actions corresponding to larger adjustment values is notably higher during the first stage than in the late stage. Conversely, during the late stage, there is a higher proportion of actions involving smaller adjustment values. It obviously aligns with the expected outcomes, validating that the agent can indeed learn a phased inversion strategy.

\subsection{Ablation Study}
\subsubsection{Contribution of State and Reward}
In this section, a series of ablation experiments are conducted to examine the effectiveness of $s_t$ and $R^t$ in our proposed RL framework. Ablation experiments present comprehensive results and comparative analysis of angle inversion outcomes, including $MAE_{angle}$, $MAEI_{angle}$, $MAE_{mean}$. We use the SD and FD to represent the temporal difference and feature difference modules of $S_t$, respectively. During the inversion process, anomalies sometimes arise during the inversion process. Thus, points with $AE_{angle}$ exceeding 50 degrees are categorized as outliers. By monitoring the number of outliers, the agent’s stability is assessed. The following table sequentially presents the quantitative results of each ablation experiment, where each set of values represents the average outcome of 1,000 experimental episodes.

\begin{table}[tbp]
\renewcommand{\arraystretch}{1.2}
\setlength{\tabcolsep}{6pt}
\caption{Ablation Study of $S_t$ and $R_t$.}
\centering
\resizebox{\linewidth}{!}{
\begin{tabular}{l|c|c|c|c}
\hline
Model & \textbf{$MAE_{\alpha}$} &\textbf{$MAE_{\beta}$}&\textbf{$MAE_{mean}$}&Outliers \\
\hline
Model with $S_{base}$ and $R_t$ & 2.159 & 3.839 & 2.999 & 9\\
Model with $S_t$ and $R_{base}$ & 2.020 & 4.452 & 3.238 & 5\\
Model with $S_t$ and $R_t$ & \textbf{1.578} & \textbf{3.662} & \textbf{2.620} & \textbf{4}\\
\hline
\end{tabular}
}
\label{Ablation_both}
\end{table}

As depicted in Table \ref{Ablation_both}, a controlled variable approach is employed to conduct three sets of experiments. The first two rows focus on ablations pertaining to the state space and reward function, respectively. Compared with the third row, it is evident that both the state space and the reward contribute to an enhancement in the inversion accuracy of both angles, thereby reducing the count of outliers. Furthermore, subsequent ablation experiments were performed on individual submodules within the state space and the reward was to further validate their effectiveness.
\subsubsection{State Components}
When the state space submodule is ablated, the reward function used is $R_{base}$. Table \ref{Ablation_St} presented that the average inversion loss of the model gradually decreases after the sequential difference module and feature difference module are gradually added to the state representation. The addition of a sequential difference module significantly improves the inversion accuracy, and the feature difference effectively reduces the number of outliers. The quantitative results highlight that the simultaneous use of the sequential difference module and the feature module effectively ensures the accuracy of the inversion of view angles.
\begin{table}[tbp]
\renewcommand{\arraystretch}{1.2}
\setlength{\tabcolsep}{6pt}
\caption{Ablation Study of $S_t$\label{tab:table4}.}
\centering
\resizebox{\linewidth}{!}{
\begin{tabular}{l|c|c|c|c}
\hline
Model & \textbf{$MAE_{\alpha}$} &\textbf{$MAE_{\beta}$}&\textbf{$MAE_{mean}$}&Outliers \\
\hline
Model with $S_t$-w/o SD, FD & 2.125 & 5.458 & 3.792 & 12\\
Model with $S_t$-w/o FD & 2.535 & \textbf{4.315} & 3.425 & 10\\
Model with $S_t$ & \textbf{2.020} & 4.452 & \textbf{3.236} & \textbf{5}\\
\hline
\end{tabular}
}
\label{Ablation_St}
\end{table}
\subsubsection{Reward Components}
To individually explain the necessity of each submodule of $R_t$, the performance of the algorithm was evaluated by sequentially removing each of the three components of the reward function. After confirming the effectiveness of the state space in Table \ref{Ablation_St}, during the ablation of the reward, the state is fixed to $S_t$. Table \ref{Ablation_Rd} presents the quantitative outcomes of various reward function configurations, which indicate that all employed reward components are effective.

\begin{table}[tbp]
\renewcommand{\arraystretch}{1.2}
\setlength{\tabcolsep}{6pt}
\renewcommand{\arraystretch}{1.2}
\setlength{\tabcolsep}{6pt}
\caption{Ablation Study of $R_t$\label{tab:table5}.}
\centering
\resizebox{\linewidth}{!}{
\begin{tabular}{l|c|c|c|c}
\hline
Model & \textbf{$MAE_{\alpha}$} &\textbf{$MAE_{\beta}$}&\textbf{$MAE_{mean}$}&Outliers \\
\hline
Model with $R_t$-w/o $R_1^t$,$R_2^t$,$R_3^t$ & 3.417 & 4.243 & 3.830 & 5\\
Model with $R_t$-w/o $R_2^t$,$R_3^t$ & 2.719 & 4.541 & 3.630 & 8\\
Model with $R_t$-w/o $R_3^t$ & 2.570 & 4.575 & 3.573 & 6\\
Model with $R_t$ & \textbf{1.578} & \textbf{3.662} & \textbf{2.620} & \textbf{4}\\
\hline
\end{tabular}
}
\label{Ablation_Rd}
\end{table}

\section{Conclusion}
%In this work, we explore how to reverse radar view angles in SAR Images using DRL. Mimicking the behavior of human beings, we propose an DRL framework, in which the SAR  agent reverse angles by interacting with an environment embedded with the electromagnetic simulator. The design of the state space effectively suppresses the background interference, enhances the agent’s sensitivity to temporal sequence changes and the ability to capture local target information. We have also designed a reward function that leverages the differential mechanism of reward memory and initial noise suppression to enhance the stability and convergence of the agent's inversion process. Through extensive experiments, we demonstrate that our proposed DRL method strikes a balance between accuracy and speed and is well-performed on both simulated and real data.

%Overall, we believe that our method is an essential stepping stone toward using RL interacting with electromagnetic simulators in solving SAR inverse problems. While our current focus has been on solving problems in lower dimensions, we plan to address various challenges in higher dimensions, including the reconstruction of 3D models from SAR images through DRL.

In this study, we investigate the application of DRL to the problem of reverse engineering radar view angles in SAR images. We develop a DRL framework that mimics human learning processes, allowing an agent to reverse-engineer angles by interacting with an environment coupled with an electromagnetic simulator. The designed state space effectively reduces background noise, enhances sensitivity to temporal changes, and improves the ability to discern local target information. Additionally, we craft a reward function that employs a differential reward memory mechanism and initial noise reduction techniques to improve the stability and convergence of the agent's learning process. Our extensive experiments validate that the proposed DRL approach achieves a commendable balance between accuracy and efficiency and performs robustly on both simulated and real-world data.

Conclusively, we posit that our method provides a foundational step towards the utilization of RL in conjunction with electromagnetic simulators to address inverse problems in SAR imagery. Although our current efforts focus on lower-dimensional challenges, future work will extend to more complex scenarios, including the construction of three-dimensional models from SAR images via advanced DRL techniques.

\end{document}